\documentclass[10pt,twocolumn,letterpaper]{article}

\usepackage{iccv}
\usepackage{times}
\usepackage{epsfig}
\usepackage{graphicx}
\usepackage{amsmath}
\usepackage{amssymb}
\usepackage{algorithm,algorithmicx,algpseudocode}
\usepackage{booktabs,xcolor,multirow,makecell}
\usepackage{import,pdfpages,bbding}
\usepackage{tikz,pgffor,pgfplots}

\usepackage[hyphens]{url}
\usepackage[pagebackref=true,breaklinks=true,letterpaper=true,colorlinks,bookmarks=false]{hyperref}

\newcommand{\bx}{\mathbf{x}}
\newcommand{\bI}{\mathbf{I}}
\newcommand{\bzero}{\mathbf{0}}

\newcommand{\bepsilon}{{\boldsymbol{\epsilon}}}

\iccvfinalcopy 

\def\iccvPaperID{****} 

\ificcvfinal\fi

\begin{document}

\title{Diffusion to Confusion: Naturalistic Adversarial Patch Generation Based on Diffusion Model for Object Detector}

\author{
    Shuo-Yen Lin\textsuperscript{1,2},
    Ernie Chu \textsuperscript{2},
    Che-Hsien Lin \textsuperscript{2},
    Jun-Cheng Chen \textsuperscript{2},
    Jia-Ching Wang \textsuperscript{1} \\
\textsuperscript{1} National Central University \\
\textsuperscript{2} Research Center for Information Technology Innovation, Academia Sinica
\\
{\tt\small \{110522139,jcw\}@cc.ncu.edu.tw;~\{shchu,ypps920080,pullpull\}@citi.sinica.edu.tw}
}

\maketitle
\ificcvfinal\thispagestyle{empty}\fi

\begin{abstract}
   Many physical adversarial patch generation methods are widely proposed to protect personal privacy from malicious monitoring using object detectors. However, they usually fail to generate satisfactory patch images in terms of both stealthiness and attack performance without making huge efforts on careful hyperparameter tuning. To address this issue, we propose a novel naturalistic adversarial patch generation method based on the diffusion models (DM). Through sampling the optimal image from the DM model pretrained upon natural images, it allows us to stably craft high-quality and naturalistic physical adversarial patches to humans without suffering from serious mode collapse problems as other deep generative models. To the best of our knowledge, we are the first to propose DM-based naturalistic adversarial patch generation for object detectors. With extensive quantitative, qualitative, and subjective experiments, the results demonstrate the effectiveness of the proposed approach to generate better-quality and more naturalistic adversarial patches while achieving acceptable attack performance than other state-of-the-art patch generation methods. We also show various generation trade-offs under different conditions. 
\end{abstract}

\section{Introduction}

In recent years, with the rapid development of deep neural networks (DNNs), various fields have been greatly benefited. In particular, computer vision has gained significant advancement in the development of object detectors. By maintaining high levels of efficiency, accuracy, and adaptability, object detection technology can be quickly applied to various scenarios such as autonomous driving~\cite{chen2015deepdriving},  quality inspection, and medical image analysis~\cite{anwar2018medical}. Although these applications bring substantial convenience to people, they also pose an increasing privacy and security threat to personal proprietary data if these technologies are abused by criminals. Besides the laws and regulations, it still has an urgent need to develop defense technologies for the public as an immediate and effective countermeasure.

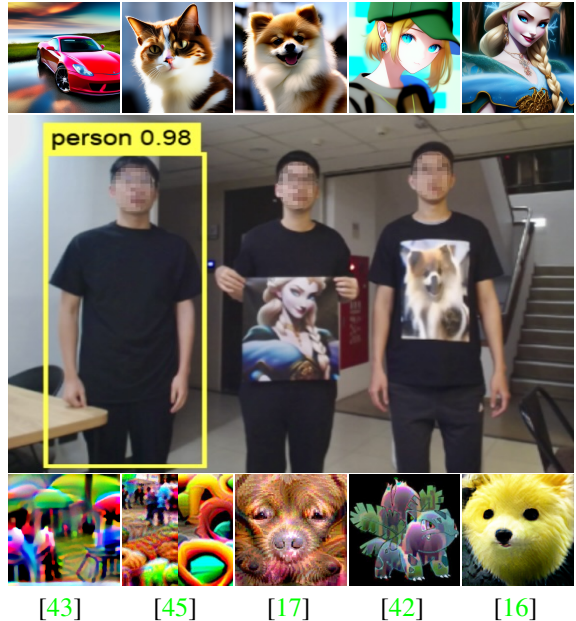
\begin{figure}[t]
\centering
\def\svgwidth{0.9\columnwidth}
\begingroup%
  \makeatletter%
  \providecommand\color[2][]{%
    \errmessage{(Inkscape) Color is used for the text in Inkscape, but the package 'color.sty' is not loaded}%
    \renewcommand\color[2][]{}%
  }%
  \providecommand\transparent[1]{%
    \errmessage{(Inkscape) Transparency is used (non-zero) for the text in Inkscape, but the package 'transparent.sty' is not loaded}%
    \renewcommand\transparent[1]{}%
  }%
  \providecommand\rotatebox[2]{#2}%
  \newcommand*\fsize{\dimexpr\f@size pt\relax}%
  \newcommand*\lineheight[1]{\fontsize{\fsize}{#1\fsize}\selectfont}%
  \ifx\svgwidth\undefined%
    \setlength{\unitlength}{564.09450715bp}%
    \ifx\svgscale\undefined%
      \relax%
    \else%
      \setlength{\unitlength}{\unitlength * \real{\svgscale}}%
    \fi%
  \else%
    \setlength{\unitlength}{\svgwidth}%
  \fi%
  \global\let\svgwidth\undefined%
  \global\let\svgscale\undefined%
  \makeatother%
  \begin{picture}(1,1.09263041)%
    \lineheight{1}%
    \setlength\tabcolsep{0pt}%
    \put(0,0){\includegraphics[width=\unitlength,page=1]{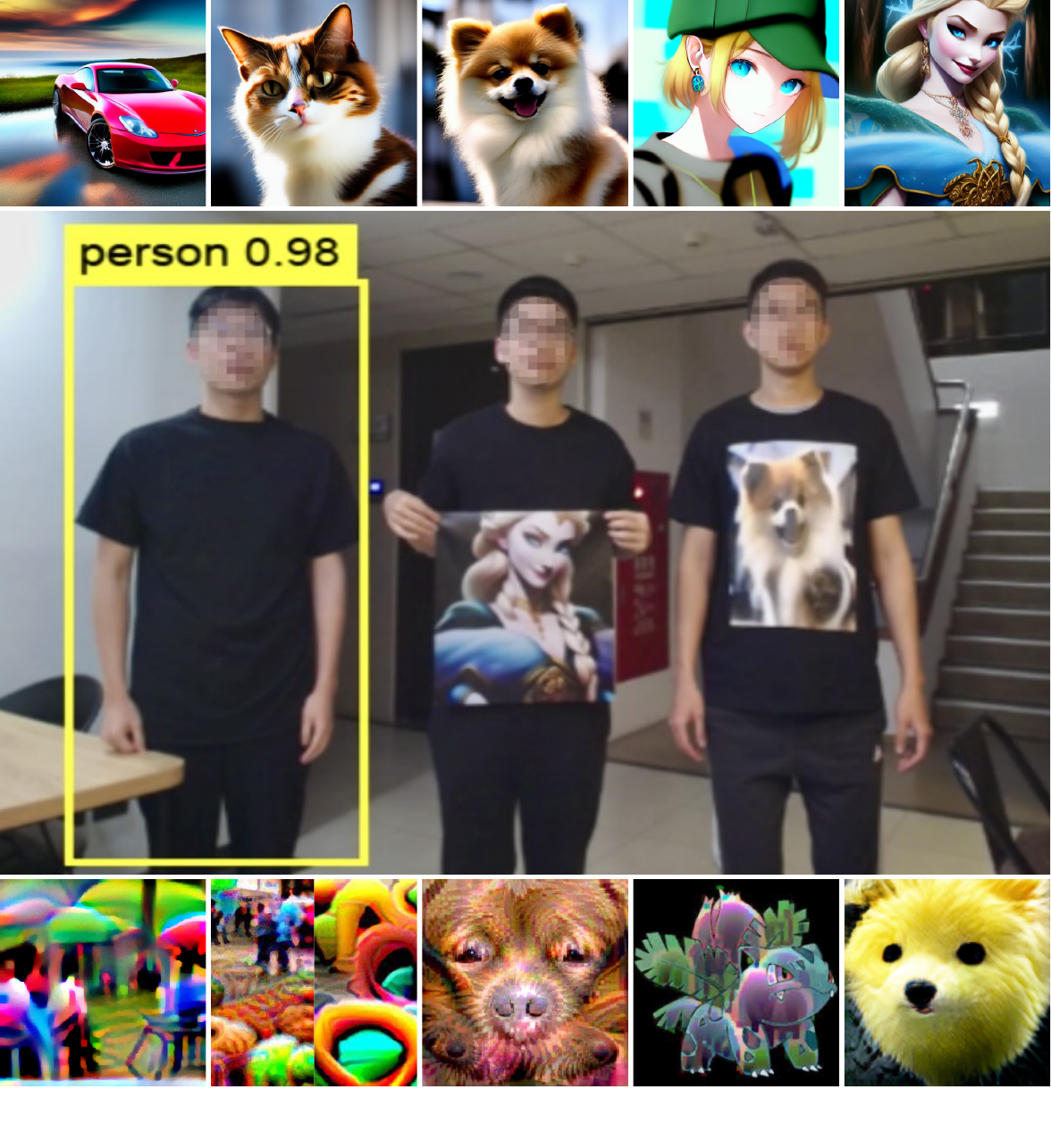}}%
    \put(0.05172425,0.00421237){\color[rgb]{0,0,0}\makebox(0,0)[lt]{\lineheight{1.25}\smash{\begin{tabular}[t]{l}\cite{thys2019fooling}\end{tabular}}}}%
    \put(0.25584103,0.00421237){\color[rgb]{0,0,0}\makebox(0,0)[lt]{\lineheight{1.25}\smash{\begin{tabular}[t]{l}\cite{makingcloak}\end{tabular}}}}%
    \put(0.65785104,0.00421237){\color[rgb]{0,0,0}\makebox(0,0)[lt]{\lineheight{1.25}\smash{\begin{tabular}[t]{l}\cite{legitimate}\end{tabular}}}}%
    \put(0.85885609,0.00421237){\color[rgb]{0,0,0}\makebox(0,0)[lt]{\lineheight{1.25}\smash{\begin{tabular}[t]{l}\cite{NPAP}\end{tabular}}}}%
    \put(0.45684604,0.00421237){\color[rgb]{0,0,0}\makebox(0,0)[lt]{\lineheight{1.25}\smash{\begin{tabular}[t]{l}\cite{UPC}\end{tabular}}}}%
  \end{picture}%
\endgroup%

\caption{Our method produces naturalistic adversarial patches that can be easily utilized in real-world. Our patches (top-row) are more diverse and natural-looking than the ones from previous works (bottom-row) .}
\label{fig:teaser}
\end{figure}

Previous works~\cite{digital_adversarial_detector2,digital_adversarial_detector1} have proposed using adversarial examples to protect people from these threats by adding small and imperceptible noise to the input data to fool DNNs to output unexpected results. 
However, since these approaches mainly consider the scenarios to attack DNNs in the digital domain, their attack success rates drop significantly when applying them physically in the real-world. To handle this challenge, several adversarial patch attacks~\cite{adversarial_stickers,adversarial_glass,thys2019fooling,adversarial_tshirt} have been proposed to achieve effective physical attacks in the real-world situations. Nevertheless, because adversarial patches only occupy a small and specific area of a whole scene, it requires significant changes to the pixel values of the patches to guarantee their attack success rates as the adversarial examples crafted using the entire image. This usually results in conspicuous and attention-grabbing patterns over the appearance of the generated patches which can not seamlessly blend into the environment and are noticed by people for their presence. Several studies~\cite{UPC,legitimate} have shown that the generated adversarial patterns can be inconspicuous and natural to people by enforcing proper constraints to preserve sufficient semantic information during patch generation. Hu \etal \cite{NPAP} further proposes to craft naturalistic adversarial patches by leveraging the learned image manifold from the pretrained generative adversarial network (GAN)~\cite{GAN} upon natural images. All these approaches can generate adversarial patches that achieve reasonable attack performance while appearing natural to humans. However, they usually require huge efforts on careful hyperparameter tuning to generate satisfactory image patches. In addition, there is still room for improvement as the quality of the generated images remain suboptimal.

On the other hand, since the emergence of diffusion models (DM) \cite{Ho_DDPM,Sohl_diffusion}, the community has significantly advanced the quality and the diversity of image generation without suffering from the mode collapse issues over other deep generative models. Although this technology has demonstrated its versatility on various applications, its effectiveness as an adversarial example generator is still unexplored. Therefore, in this work, we present a novel naturalistic adversarial patch generation method based on DM. In addition, we also provide rich hands-on experience to strike a balance between leveraging the pretrained DM upon natural images to generate high-fidelity and naturalistic patches to humans and maintaining sufficient attack capability. With extensive experiments, the results demonstrate the effectiveness of the proposed approach to generate better-quality and more naturalistic adversarial patches than other similar patch generation methods as shown in Figure~\ref{fig:teaser}. Moreover, the proposed method can adapt to any detector to effectively decrease its performance in detecting humans.

Our main contributions are two-fold. First, to the best of our knowledge, we are the first to propose DM-based naturalistic adversarial patch generation for object detectors. Next, we present rich hands-on experience to show how to generate visually pleasing and naturalistic patches to humans while achieving comparable attack performance. The extensive experiments and analyses are conducted to show different generation trade-offs under different conditions.

\section{Related work}\label{sec:related}
In this section, we provide a brief overview of recent relevant works covering different adversarial attacks, including digital and physical attacks. An adversarial attack is a technique used to create adversarial examples that can deceive target models, such as classification or detection models, to produce inaccurate outputs. In general, adversarial examples can be categorized into two types: digital and physical adversarial examples.

\subsection{Digital adversarial example}
It denotes that the victim model is attacked by directly taking the digital images manipulated with imperceptible adversarial perturbation as inputs and generating wrong predictions. Szegedy \etal \cite{szegedy} pioneered to show that the imperceptible adversarial examples can be easily crafted by solving an optimization problem to enforcing the victim model to output incorrect results using the L-BFGS algorithm with box constraints. Similarly, Carlini \etal \cite{carlini_attack} also show another way to obtain adversarial examples by solving similar optimization problem with different distance metrics using the Adam optimizer \cite{Adam}. Other popular and common methods to generate adversarial examples include Fast Gradient Sign Method (FGSM)~\cite{FGSM}, which computes the gradient of the loss function with respect to the input and perturbs the input in the direction of the sign of the computed gradient multiplied by a small constant. Moreover, Iterative-FGSM (I-FGSM)~\cite{IFGSM} and Projected Gradient Descent (PGD)~\cite{PGD} in general employ multiple iterations of FGSM to obtain stronger adversarial examples.

\subsection{Physical adversarial example}
It denotes that the adversarial examples are printed out and recaptured by a camera. Kurakin \etal \cite{IFGSM} show the printed adversarial images not only affect the victim models in the digital domain but can also persist in the physical world. In addition, Athalye \etal~\cite{EOTAthalye} further propose Expectation Over Transformations (EOT) strategy to simulate the possible transformations (\eg, rotation, scaling, brightness changes, \etc.) that the adversarial examples could encounter in the physical world to enhance their physical attack capability. Following this research direction, different physical adversarial attacks are developed for various tasks, such as object classification and detection. For instance, Brown \etal~\cite{adversarialpatch_first} develop an adversarial patch generation method where the generated patch image can be placed on an object and fool a well-trained classifier to yield a wrong class prediction. Similarly, in the field of object detection, researchers have shown various ways to alter the real-world road signs to craft physical adversarial examples by using camouflage graffiti or pasting stickers on them as shown in \cite{stopsign4,stopsign1,stopsign_sticker,stopsign2,stopsign3}, in which they can successfully deceive the model to regard the road signs as other objects. Besides traffic signs, other works have explored on accessories that people wear, such as hat \cite{adversarial_hat}, glasses \cite{adversarial_glass}, masks \cite{adversarial_stickers_face_reco} and clothing \cite{NPAP,UPC,legitimate,thys2019fooling,makingcloak,adversarial_tshirt}.
\algrenewcommand\algorithmicindent{1em}%
\begin{figure*}[t]
\begin{minipage}[t]{0.42\textwidth}
\begin{algorithm}[H]
  \caption{Diffusion Model sampling} \label{alg:ddpm-sampling}
  \small
  \begin{algorithmic}[1]
    \vspace{.04in}
    \State $\bx_T \sim \mathcal{N}(\bzero, \bI)$
    \For{$t=T, \dotsc, 1$}
      \State $ z \sim \mathcal{N}(\bzero, \bI)$ if $t > 1$, else $z = \bzero$
      \State $\bx_{t-1} = \frac{1}{\sqrt{{\alpha_t}}} \left(\bx_t-\frac{1-\alpha_t}{\sqrt{1-\Bar{\alpha_t}}}\bepsilon_\theta(\bx_t, t) \right)+\Sigma_tz$
    \EndFor
    \State \textbf{return} $\bx_0$
    \vspace{.24in}
  \end{algorithmic}
\end{algorithm}
\end{minipage}
\hfill
\begin{minipage}[t]{0.56\textwidth}
\begin{algorithm}[H]
  \caption{Adversarial patch sampling (APS)} \label{alg:adv-sampling}
  \small
  \begin{algorithmic}[1]
    \vspace{.04in}
    \State \textbf{Input:} Patch $x_0$, Starting time step $t_{start}$, Step size $s$, Condition $c$
      \State $t=t_{\textit{start}}$; \; $\bx_t = \sqrt{\Bar{\alpha}_t} \bx_0 + \sqrt{1-\Bar{\alpha}_t}z$; \; $z \sim \mathcal{N}(\bzero, \mathbf{I})$
      \Repeat
        \State $x_{t-1}=\sqrt{\Bar{\alpha}_{t-1}} \left(\frac{x_{t}-\sqrt{1-\Bar{\alpha}_{t}} \cdot\epsilon_{\theta}\left(x_{t},t,c\right)}{\sqrt{\Bar{\alpha}_{t}}}\right)+\sqrt{1-\Bar{\alpha}_{t-1}} \cdot \epsilon_{\theta}\left(x_{t},t,c\right)$

        \State $t=t-s$
      \Until{$t < 2s$}
      \State \textbf{return} $\sqrt{\Bar{\alpha}_{0}} \left(\frac{x_{t}-\sqrt{1-\Bar{\alpha}_{t}} \cdot\epsilon_{\theta}\left(x_{t},t,c\right)}{\sqrt{\Bar{\alpha}_{t}}}\right)+\sqrt{1-\Bar{\alpha}_{0}} \cdot \epsilon_{\theta}\left(x_{t},t,c\right)$
    \vspace{.04in}
  \end{algorithmic}
\end{algorithm}
\end{minipage}
\end{figure*}

\subsection{Naturalistic adversarial patch}
On the other hand, since adversarial patches only occupy a small and specific area of a whole scene when performing attack, it requires much stronger changes to the pixel values of the patches to have similar attack strength than the adversarial examples crafted using the entire image. To avoid making the adversarial patch look unnatural and attract attention, some works~\cite{stopsign_natural,stopsign_natural2} have proposed different constraints to preserve a natural appearance for the generated adversarial examples while maintaining the adversarial attack capacity. Furthermore, Tan \etal~\cite{legitimate} propose an effective measure to quantify the legitimacy of the adversarial patch by examining indicators of rationality, such as color, edge, and texture features. Then, they leverage this measure to constrain the adversarial perturbation from being too noticeable or obvious. In contrast to above methods, Hu \etal~\cite{NPAP} propose to craft naturalistic adversarial patches using GAN. By leveraging the learned image manifold from the pretrained GAN upon natural images, the generated adversarial patches achieve satisfactory attack performance while appearing natural to humans.

To sum up, \cite{NPAP} is the most relevant work to the proposed method which the generated adversarial patch can be pasted on cloth to fool object detectors while also appearing natural to people. As compared with them, since the proposed method exploits the pretrained diffusion model on natural images which allows to synthesize higher-quality and more diverse patch images than GAN without mode collapse issues, our method strikes a better balance between richness and naturalness than \cite{NPAP} to produce adversarial patches that are both more inconspicuous and more diverse.

\vspace{-0.3cm}
\section{Background}
To better understand our method, it is helpful to have a general concept of Diffusion Models (DM) \cite{Ho_DDPM,Sohl_diffusion}. DM is a latent variable model that comprises two parts: a forward time diffusion process and a backward time generative model (Sec. \ref{sec:dm}). In addition to the original DM, several enhancements have been proposed to significantly reduce the computation (Sec. \ref{sec:ddim}) and memory usage (Sec. \ref{sec:ldm}), which make our approach feasible.

\subsection{Diffusion models}
\label{sec:dm}
During the forward time process, progressive Gaussian noise is added to the given data points from a real data distribution $\bx_0 \sim q(x)$, resulting in a latent sequence of $T$ elements, $[\bx_1, \dots, \bx_T]$, that goes from clean to noisy. The distribution of latent variable $\bx_t$ conditioned on $\bx_0$, for any $t \in \mathbb{Z}, 1 \leq t \leq T$ is given by
\begin{equation}
q(\bx_t | \bx_0) = \mathcal{N}(\bx_t; \sqrt{\Bar{\alpha}_t} \bx_0, (1-\Bar{\alpha}_t)\mathbf{I}),
\label{eq:diffuse}
\end{equation}
where $\Bar{\alpha}_t = \Pi_{i=1}^t \alpha_i$ is a well-designed strictly negative function of $t$ \cite{VDM}, and $\alpha_t$ is the hyperparameter that controls noise scheduling. As $t$ increases, $\bx_t$ gradually loses its original data features and becomes pure noise from a standard normal distribution when $t=T$, i.e., $\bx_T \sim \mathcal{N}(\bzero, \mathbf{I})$.

In the backward time generative model, the goal is to reverse the aforementioned process and generate realistic samples from pure noises. The synthetic distribution $p_{\bepsilon_\theta}(\bx_{0:T}) = p(\bx_T)\prod_{t=1}^T p_\theta(\bx_{t-1}|\bx_t)$, parameterized by a neural noise predictor $\bepsilon_\theta$, is optimized to approximate the data distributions at different time steps.


In practice, the noise predictor $\bepsilon_\theta(\bx_t, t)$ infers the residual noise from $\bx_t$ and thus gives $\hat{\bx}_0$ (estimated clean data samples), which is accurate enough to construct a less noisy distribution $q(\bx_{t-1} |\bx_t ,\hat{\bx}_0)$. So far, one can sample a data point from the marginal distribution as shown in Algorithm \eqref{alg:ddpm-sampling}. While the previously described approach yields impressive sample quality, it can be computationally expensive to run iterative inference for a single data point.

\subsection{Speed up generation process}
\label{sec:ddim}
Song \textit{et al.} \cite{Song_ddim} proposed DDIM, which substitutes the noise term in $\bx_{t-1} \sim q(\bx_{t-1}|\bx_0)$ with $\bx_t$ and introduces a randomness factor $ \sigma_t$, derived by:
\begin{equation}
\begin{aligned}
\bx_{t-1} &\sim q(\bx_{t-1} | \bx_0), \quad \bepsilon_{t-1}, \bepsilon_t, \bepsilon \sim \mathcal{N}(0, \mathbf{I}) \\
\bx_{t-1} &= \sqrt{\Bar{\alpha}_{t-1}} \bx_0 + \sqrt{1-\Bar{\alpha}_{t-1}}\bepsilon_{t-1} \\
\bx_{t-1} &= \sqrt{\Bar{\alpha}_{t-1}} \bx_0 + \sqrt{1-\Bar{\alpha}_{t-1} - {\sigma}^2_t} \bepsilon_{t} + {\sigma}_t \bepsilon \\
\end{aligned}
\label{eq:ddim}
\end{equation}
The magnitude of ${\sigma}_t$ determines the level of stochasticity in the forward process is. As ${\sigma}_t$ decreases towards zero, the diffusion process becomes less stochastic and no longer remains Markovian, as $\bx_t$ depends not only on $\bx_0$ but also on $\bx_{t+1}$. This non-Markovian characteristic enables sampling with fewer time steps, which reduces the computation barrier.

\subsection{Diffusion on low-dimensional latent space} 
\label{sec:ldm}
In order to further improve computation efficiency, Latent Diffusion Model (LDM) \cite{LDM} utilizes an autoencoder to map images into a low-dimensional latent space that is perceptually equivalent to the image space. This enables the diffusion process to operate on a more manageable dimensionality. Moreover, external information such as semantic embeddings from a language model can be integrated to perform conditional generation. Conditions $c$ can be injected into the noise predictor $\bepsilon_\theta(x_t, t, c)$ using either cross-attention or concatenation.

Thus far, DM has exhibited superior properties compared to various generative models, without imposing excessive computation burdens. This makes it a perfect candidate as an adversarial patch generator.

\section{The proposed method}
\begin{figure}
\centering
\def\svgwidth{\columnwidth}
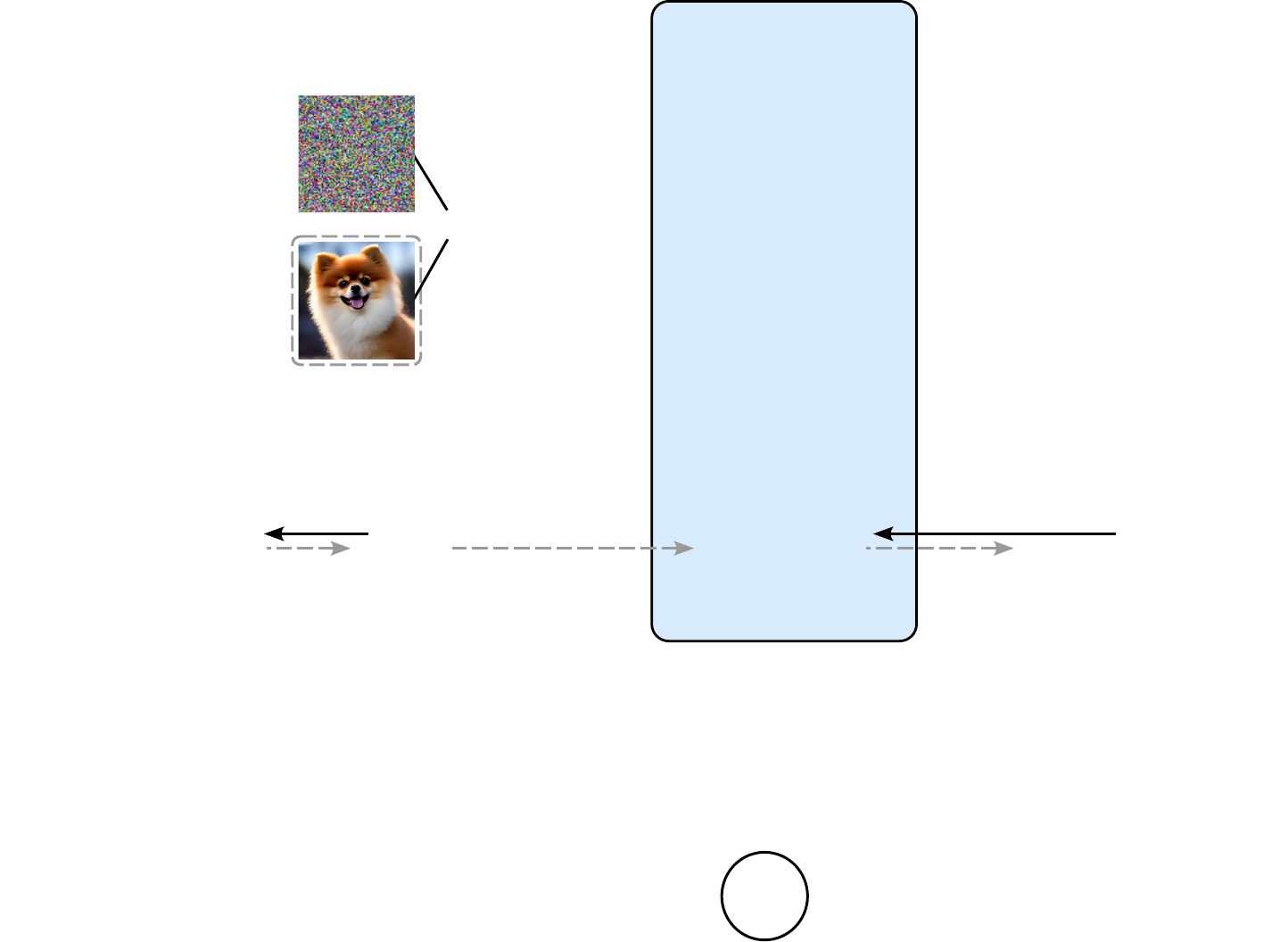
\caption{Overall architecture for the proposed method which generates naturalistic adversarial patches. At each optimization iteration, a fixed amount of Gaussian noise is added to the patch. A denoising U-Net then iteratively removes the noise from the patch. Finally, we update the original patch according to the gradient from the object detector's objective. Human image courtesy to INRIA dataset.}
\label{fig:system-diagram}
\end{figure}

In order to obtain naturalistic adversarial patches that deceive the object detector, we first generate an initial patch using a pretrained text-conditioned LDM, and apply the patch to the training data using the scene rendering function (Sec. \ref{sec:init-patch}). Rather than directly parameterizing the objective of the victim object detector on the image patch, we back-propagate further into the sampling process of LDM and update the internal latent with SGD (Sec. \ref{sec:loss}). Lastly, we show how these designs help maintaining the naturalness of the patch (Sec. \ref{sec:maintain-natural}). The overall framework is depicted in Figure \ref{fig:system-diagram}. 

\subsection{Naturalistic adversarial patch generation}
\label{sec:init-patch}
We begin with how to produce our patch and apply it to the pedestrian training data.
\paragraph{Patch generation.}
To generate the adversarial patch, we introduce the Adversarial Patch Sampling (APS) method in Algorithm \eqref{alg:adv-sampling}. Derived from Equation \eqref{eq:ddim} with the randomness factor $\Bar{\sigma}_t=0$, APS employs a pretrained text-conditioned LDM to generate a naturalistic patch $P$ with an arbitrary text description $c_P$. The generation process of the initial patch can be formulated as:
\begin{equation}
  P_{\textit{init}} = \text{APS}(\texttt{NULL}, T, s, c_P),
\end{equation}
where the non-Markovian step size $s > 1$ accelerates the sampling process. Since the starting time step is $T$, $P_{\textit{init}}$ is denoised from a pure noise and does not depend on the first argument of APS.

\paragraph{Scene rendering.}
To simulate the scene where the patch $P$ is pasted on a person, we use a scene rendering function $\mathcal{R}$. Given a pedestrian scene image $\mathcal{I}$, we can obtain the rendered scene image $\mathcal{I}_P$ by:
\begin{equation}
    \mathcal{I}_P=\mathcal{R}(\mathcal{I},\mathcal{A}_\phi(P)),
\end{equation}
where $\mathcal{A_\phi}$, parameterized by a random variable $\phi$, is an image transformation that simulates the physical conditions such as illumination, contrast and noise. $\mathcal{R}$ utilizes the victim object detector to find the people in the scene and renders the adversarial patch on the desired location, as shown in Figure \ref{fig:system-diagram}.

\subsection{Loss function}
\label{sec:loss}

With the sampling method APS and the rendered scene $\mathcal{I}_P$, to turn the initial patch $P = P_{\textit{init}}$ into an adversarial patch that can evade object detectors, we define the following objective function for this problem:
\begin{equation}
    \min_{P} \left( \frac{1}{N} \sum^N_{i=1} \mathcal{L}_{det}(\mathcal{I}^i_{P'})+\lambda\cdot \mathcal{L}_{tv}(P') \right),
    \label{eq:loss-fn}
\end{equation}
where $P' = \text{APS}(P, t_{\textit{start}}, s, c_P)$ is the resampled patch, and $N$ is the number of images in a minibatch. By using this objective function, we can obtain the adversarial gradient and use it to guide the generation of the adversarial patch.

The detector loss $\mathcal{L}_{det}$~\cite{thys2019fooling} in the main objective aims to reduce the detection confidence of the most salient person in $\mathcal{I}_P$. The loss can be expressed as follows:
\begin{equation}
\mathcal{L}_{det}(\mathcal{I}_P) = \max_j \left(\mathcal{D}^{j}_{obj}(\mathcal{I}_P)\mathcal{D}^{j}_{cls_k}(\mathcal{I}_P)\right),
\end{equation}
where $\mathcal{D}^{j}_{obj}$ is the objectness score of the $j$th bounding box, and $\mathcal{D}^{j}_{cls_k}$ is the probability that the $j$th bounding box belongs to class $k$. We only focus on the case where $k$ is \texttt{Person}.


In addition to the detection score, we also use the total variation loss, denoted as $L_{tv}$. This regularization term promotes smoothness in the generated images by penalizing the total variation between adjacent pixels in the image. 
Let $p_{x,y} \in P$ be the image pixel of the patch, we have:
\begin{equation}
\mathcal{L}_{tv}(P) = \sum_{x,y}\sqrt{(p_{x+1,y}-p_{x,y})^2+(p_{x,y+1}-p_{x,y})^2}.
\end{equation}


\subsection{Maintaining naturalness}
\label{sec:maintain-natural}

In this section, we delve deeper into the details of our method and explain the components we choose to promote the naturalness of adversarial patches.
\vspace{-0.3cm}
\paragraph{Using text condition.} 
 
It is easy for $P$ to lose its natural appearance before we find the optimal $P$ that best satisfies the objective. To find a natural-looking patch that possesses adversarial capabilities to a certain extent, we use the text conditioning to guide the patch generation. In our ablation studies, we show that using a strong conditioning like text is crucial to maintain the naturalism of the patch during the search of adversaries.

In addition, the reason we use generated images as the naturalistic patch instead of arbitrary real-world images is that the generated image has a better alignment with the conditioning text description \wrt the pretrained LDM, and the distribution of a random real-world image is not guaranteed to be well-captured by the pretrained LDM.

\vspace{-0.3cm}

\paragraph{Search strategy.} 
Previous methods directly apply the patch $P$ on the training data $\mathcal{I}$, \ie let $P' = P$. In contrast, our approach introduces more diversity while maintaining its semantic content and natural appearance by Equation~\eqref{eq:loss-fn}. Specifically, we use the diffusion process described in Equation~\eqref{eq:diffuse} to add noise into the patch $P$ to a specific level $t_{\text{start}}$. This step introduces randomness into the patch. Our ablation studies empirically demonstrate that the randomness brings up more candidate optimizing directions that promote attack ability while maintaining image naturalness. Next, we perform the denoising process described in Algorithm \eqref{alg:adv-sampling}, using the same condition $c_P$ as that used for the initial patch. We can thus obtain the resampled patch $P'$ and apply it to the training data to create $\mathcal{I}_{P'}$.    

\begin{table*}
\centering
\caption{Cross-model attack performance (mAP\%, lower is better). Best results are \underline{underlined}, and top-2 best results are in \textbf{bold}.}
\label{tab:cross-model}
\begin{tabular*}{\linewidth}{@{\extracolsep{\fill}} llcccccccccc}
\toprule
\multicolumn{2}{l}{Method} & YL2 & YL3 & YL3t & YL4 & YL4t & FRCNN & DDETR & YL5s & YL7t & \textbf{Avg.} \\
\midrule
\multicolumn{2}{l}{{\scriptsize $(P_T)$} Thys \etal \cite{thys2019fooling} $^*$} & \textbf{\underline{2.68}}  & \textbf{\underline{22.51}} & \textbf{8.74}  & \textbf{\underline{12.89}} & \textbf{\underline{3.25}}  & \textbf{39.41} & \textbf{34.46} & \textbf{34.00} & \textbf{\underline{20.81}} &  \textbf{\underline{19.86}}\\
\multicolumn{2}{l}{{\scriptsize $(P_H)$} Hu \etal \cite{NPAP} $^\dagger$} & 34.76  & 37.79 & 21.69  & 46.80 & 8.67  & 59.97 & 55.85 & 38.19 & 29.20 & 36.99 \\
\midrule
\multirow{12}{*}{\rotatebox{90}{Ours (trained on)}} & {\scriptsize $(P_1)$} YL2 & \textbf{11.62} & 35.86 & 31.55 & 60.25 & 29.94 & 52.02 & 52.37 & 57.97 & 39.00 & 41.18\\
&{\scriptsize $(P_2)$} YL3       & 23.03 & \textbf{28.53} & 26.25 & 59.96 & 36.05 & 53.76 & 56.23 & 58.04 & 36.15 & 42.00\\
&{\scriptsize $(P_3)$} YL3t   & 34.18 & 38.52 & \textbf{\underline{8.31}}  & 65.20 & 31.74 & 58.15 & 57.23 & 57.07 & 33.37 & 42.64 \\
&{\scriptsize $(P_4)$} YL4       & 53.50 & 67.10 & 51.85 & \textbf{14.47} & 60.97 & 60.40 & 62.40 & 68.16 & 58.70 & 55.28\\
&{\scriptsize $(P_5)$} YL4t   & 22.80 & 34.69 & 22.54 & 68.89 & \textbf{8.45}  & 51.26 & 52.51 & 53.37 & 30.61 & 38.35\\
&{\scriptsize $(P_6)$} FRCNN   & 30.14 & 53.58 & 36.34 & 61.99 & 42.43 & \textbf{\underline{37.81}} & 35.18 & 59.94 & 36.65 & 43.78 \\
&{\scriptsize $(P_7)$} DDETR       & 28.97 & 55.99 & 49.78 & 48.63 & 47.18 & 48.82 & \textbf{\underline{22.04}} & 62.69 & 42.30 & 45.16 \\
&{\scriptsize $(P_8)$} YL5s      & 25.85 & 40.92 & 20.32 & 60.09 & 35.95 & 56.66 & 46.32 & \textbf{\underline{33.26}} & 30.55 & 38.88 \\
&{\scriptsize $(P_9)$} YL7t   & 28.78 & 41.63 & 25.70 & 59.52 & 33.77 & 52.42 & 58.09 & 53.44 & \textbf{25.32} & 42.07 \\
\cmidrule(lr){2-12}
&{\scriptsize $(P_{10})$} Ensemble & 15.46 & 30.86 & 17.96 & 44.45 & 17.41 & 49.31 & 40.93 & 47.91 & 26.83  & \textbf{32.35}\\
\midrule\midrule
\multicolumn{2}{l}{{\scriptsize $(P_{A1})$} Random Noise} & 75.46 & 74.49 & 79.66 & 77.40 & 78.15 & 72.66 & 73.49 & 74.92 & 48.81 & 72.78 \\
\multicolumn{2}{l}{{\scriptsize $(P_{A2})$} Unoptimized Patch} & 62.74 & 71.15 & 65.60 & 63.55 & 63.84 & 67.01 & 68.49 & 74.38 & 59.10 & 66.21 \\
\bottomrule
\multicolumn{12}{c}{\small $^*$Trained on YL2. $^\dagger$Trained on YL4t. All results are the best available.}
\end{tabular*}

{\setlength{\arrayrulewidth}{1pt}
\begin{tabular}{c|c}
\foreach \n in {1,3,5,7,9}{
  \begin{tabular}{@{}c@{}}
    \includegraphics[width=0.124\textwidth]{figs/cross-model/P\n.jpg} \\
    {\footnotesize $P_{\n}$}
  \end{tabular}
}
&
\foreach \n in {T,A1}{
  \begin{tabular}{@{}c@{}}
    \includegraphics[width=0.124\textwidth]{figs/cross-model/P\n.jpg} \\
    {\footnotesize $P_{\n}$}
  \end{tabular}
} \\
\foreach \n in {2,4,6,8,10}{
  \begin{tabular}{@{}c@{}}
    \includegraphics[width=0.124\textwidth]{figs/cross-model/P\n.jpg} \\
    {\footnotesize $P_{\n}$}
  \end{tabular}
}
&
\foreach \n in {H,A2}{
  \begin{tabular}{@{}c@{}}
    \includegraphics[width=0.124\textwidth]{figs/cross-model/P\n.jpg} \\
    {\footnotesize $P_{\n}$}
  \end{tabular}
}
\end{tabular}}
\end{table*}
\section{Experiments}

\paragraph{Implementation details.}
In our experiments, we use the modified Stable Diffusion \cite{LDM} as the pretrained text-conditioned LDM. For the most of our results, we set the total number of diffusion steps $T$ to 1,000 and the starting time step $t_{\textit{start}}$ to 500 (corresponding to halfway noise level in APS. To achieve our goal of 3 denoising iterations in the optimization process, we rounded down the step size $s$ to 166. Unless stated otherwise, we use this setting for all of our experiments. The patch is optimized using the Adam optimizer, with an initial learning rate of 0.005 and decay if the change in losses is consistently below $1e^{-4}$. The training is performed on a single NVIDIA GeForce RTX 3090.

\vspace{-0.2cm}
\paragraph{Victim models.}
In addition to the model families previously attacked, including YOLOv2 \cite{yolov2}, YOLOv3 \cite{yolov3}, YOLOv4 \cite{yolov4} and FasterRCNN \cite{fasterRCNN}, we also evaluate recent state-of-the-art models, such as Deformable DETR \cite{D_Detr}, YOLOv5 \cite{yolov5} and YOLOv7 \cite{yolov7}. All of these models have been trained on the COCO dataset \cite{MSCOCO}, which comprises 80 classes, including the \texttt{Person} class. We use abbreviations for YOLOv as YL, FasterRCNN as FRCNN, and Deformable DETR as DDETR in our tables. To differentiate between the various variants of YOLO models, we use the letter `t' to denote the tiny version and `s' to denote the small version. For example, YL2t refers to YOLOv2-tiny, and YL5s refers to YOLOv5-small. For all object detection models, the images are resized to 416x416 for both training and evaluation.

\setlength\tabcolsep{0pt}
\begin{table}
\centering
\caption{Attack performance (mAP\%) and subjective preference. Best results are in \textbf{bold}.}
\label{tab:inria}
\begin{tabular*}{\linewidth}{@{\extracolsep{\fill}} lccccccc}
\toprule
Method & YL2 & YL3 & YL3t & YL4  & YL4t & FRCNN & \textbf{Avg.} \\
\midrule
Hu \etal \cite{NPAP} & 12.1  & 34.9  & 10.0    & 22.6  & 8.7    & 42.5  & 21.8\\
Ours    & \textbf{11.6}  & \textbf{28.5}  & \textbf{8.3}     & \textbf{14.5}   & \textbf{8.5}    & \textbf{37.8}  & \textbf{18.2}\\ 
\midrule
\makecell{\color{blue} \% of users \\\color{blue} prefer ours} & \color{blue} 85.1 & \color{blue} 92.1 & \color{blue} \color{blue} 64.0 & \color{blue} 93.0 & \color{blue} 86.8 & \color{blue} 64.0 & \color{blue} 80.8 \\
\bottomrule
\end{tabular*}
\end{table}
\vskip -0.2cm

\vspace{-0.2cm}
\paragraph{Dataset.}
The INRIA Person dataset \cite{INRIA} is predominantly used in our experiments since it is a widely-used dataset for pedestrian detection tasks. The dataset contains 614 and 288 person images for the training and testing set, respectively. To enable cross-dataset evaluation, we use the same testing split from Hu \etal \cite{NPAP} for the MPII dataset \cite{MPII}. 

\vspace{-0.2cm}
\paragraph{Evaluation setup.}
Following the approach proposed in \cite{NPAP}, we utilize the object detector's predictions on clean test data (without applying any patch) as the reference label. We measure the success of the attack by calculating the mean average precision (mAP) of the complete test dataset. In such scenario, we standardize the detectors' performance on clean images to an mAP of 100\%, allowing for a straightforward comparison of the performance degradation.

\begin{table}
\centering
\caption{\label{tab:naturalness_test} Subjective naturalness evaluation (The score for each test image is calculated by the ratio of positive votes, higher is better).}
\begin{tabular*}{\linewidth}{@{\extracolsep{\fill}} ccccc}
    \toprule
    \makecell{Thys \etal \\\cite{thys2019fooling}}
    & \makecell{Tan \etal \\\cite{legitimate}}
    & \makecell{Hu \etal \\\cite{NPAP}}
    & {Ours}
    & \makecell{Real-world \\image}\\
    \midrule
    \includegraphics[width=0.19\linewidth]{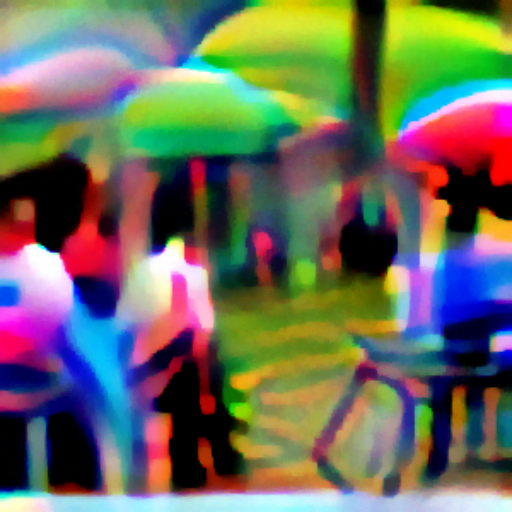}
    & \includegraphics[width=0.19\linewidth]{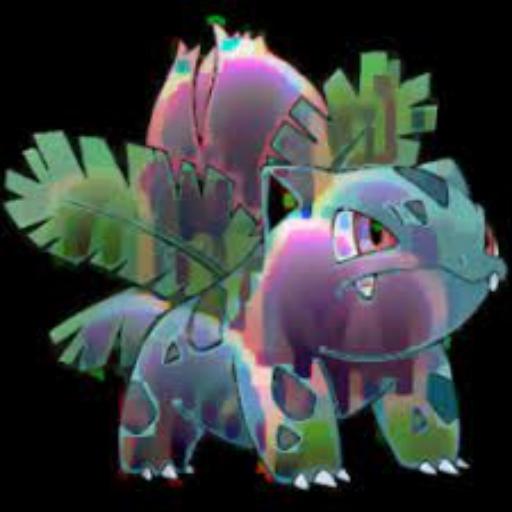}
    & \includegraphics[width=0.19\linewidth]{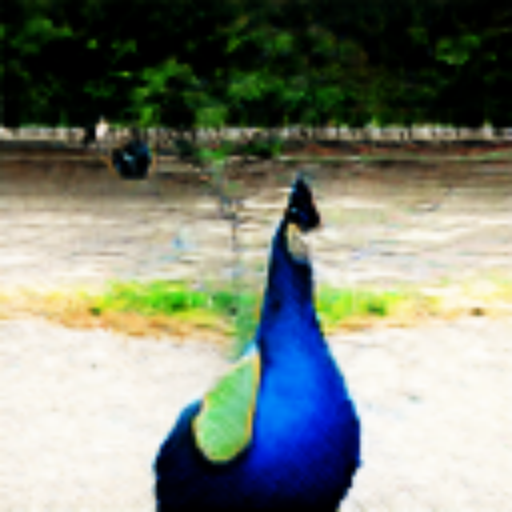}
    & \includegraphics[width=0.19\linewidth]{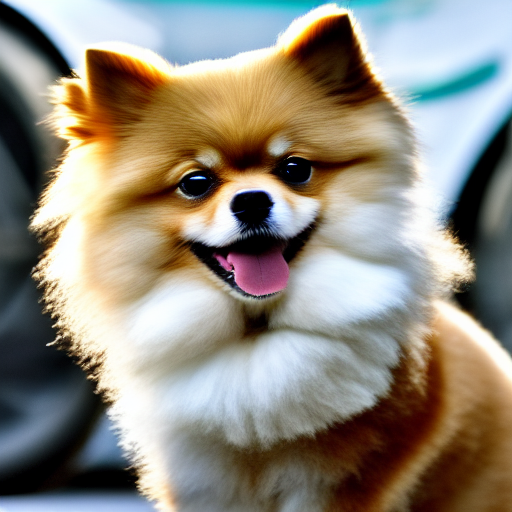}
    & \includegraphics[width=0.19\linewidth]{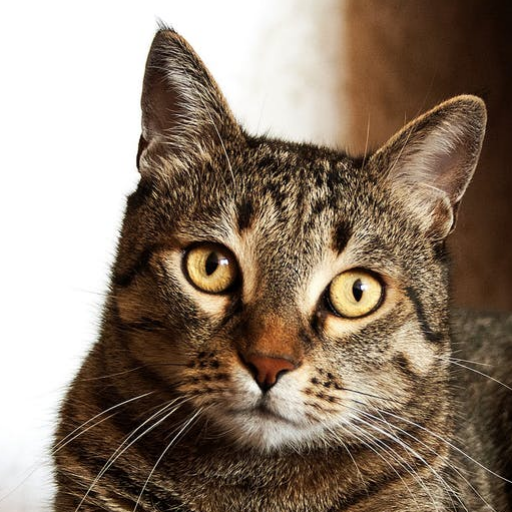}\\
    {2.63\%}
    & {0.01\%}
    & {14.03\%}
    & {69.29\%}
    & {99.99\%}\\
\bottomrule
\end{tabular*}
\end{table}

\begin{table*}
\centering
    \caption{\label{tab:cross-dataset}Cross-dataset attack performance (mAP\%) using YL4t. Best results are in \textbf{bold}.}
    \begin{tabular*}{\linewidth}{@{\extracolsep{\fill}} lcccccccccc}
        \toprule
        \multirow{2}{*}{Method} & \small Trained on & \multicolumn{3}{c}{INRIA} & \multicolumn{3}{c}{MPII} & \multicolumn{3}{c}{Mix} \\
        \cmidrule(){2-2}\cmidrule(lr){3-5} \cmidrule(lr){6-8} \cmidrule(lr){9-11}
        & \small Tested on & INRIA & MPII & Mix & INRIA & MPII & Mix & INRIA & MPII & Mix \\
        \midrule
        \multicolumn{2}{l}{Hu \etal \cite{NPAP}} & 8.67 & \textbf{0.51} & \textbf{2.69} & 22.05 & 7.92 & 14.12 & 18.45 & 6.32 & 11.68 \\
        \multicolumn{2}{l}{Ours} & \textbf{8.45} & 2.38 & 4.99 & \textbf{15.59} & \textbf{6.10} & \textbf{10.24} & \textbf{12.38} & \textbf{4.04} & \textbf{7.64} \\
        \bottomrule
    \end{tabular*}
\end{table*}

\begin{table*}
\centering
\caption{\label{tab:physical_attack} Attack performance (mAP\%) in different physical environment using YL4t. Results on previous methods are annotated with $^*$.}
\begin{tabular*}{\linewidth}{@{\extracolsep{\fill}} ccccccc}
    \toprule
    
    \multicolumn{2}{c}{Indoor scenes} & \multicolumn{2}{c}{Outdoor scenes}& \multicolumn{2}{c}{Compare to Hu \etal \cite{NPAP} $^*$} \\
    \cmidrule(lr){1-2}\cmidrule(lr){3-4}\cmidrule(lr){5-6}
    
     \includegraphics[width=0.16\linewidth]{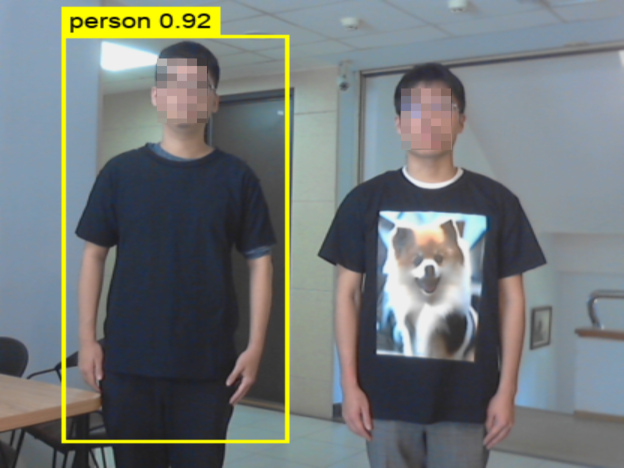}
    & \includegraphics[width=0.16\linewidth]{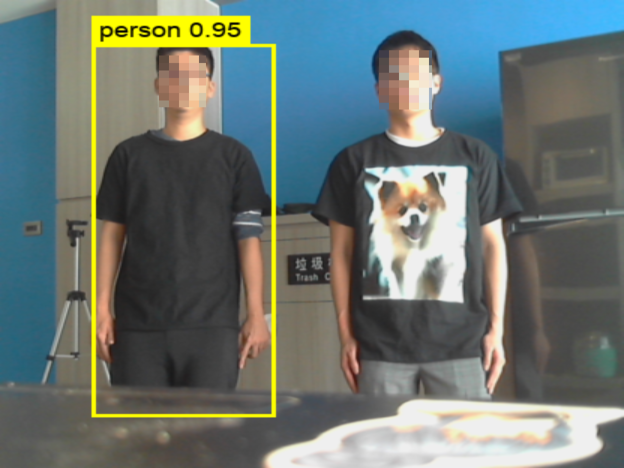}
    & \includegraphics[width=0.16\linewidth]{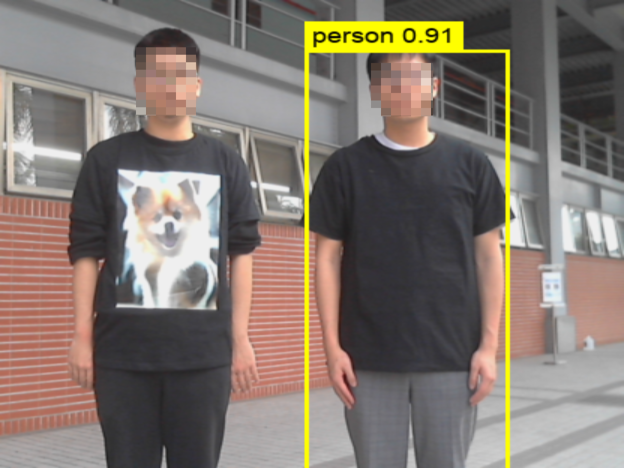}
    & \includegraphics[width=0.16\linewidth]{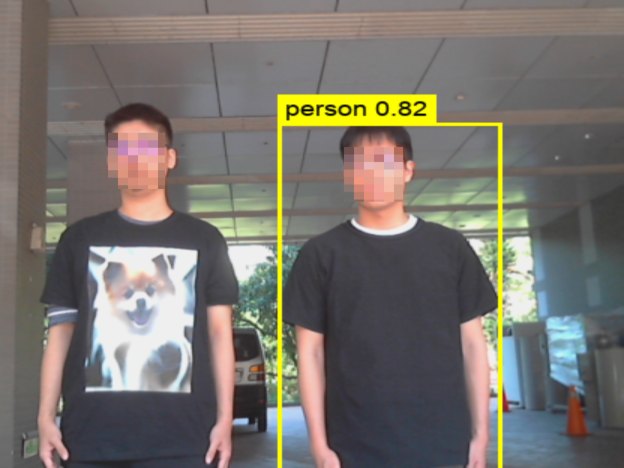}
    & \includegraphics[width=0.16\linewidth]{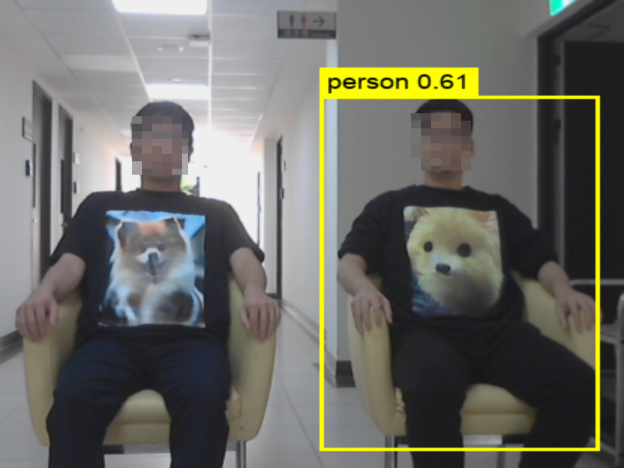} 
    & \includegraphics[width=0.16\linewidth]{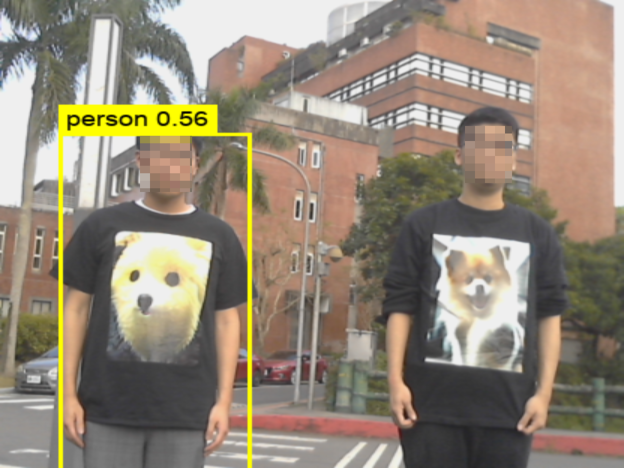} \\
    \midrule
    23.33 & 28.80 & 22.72 & 24.50 & 21.21 $|$ $^*$25.00 & $^*$33.21 $|$ 27.45  \\
    \bottomrule
\end{tabular*}
\end{table*}

\subsection{Cross-model generalizability}
We train our naturalistic adversarial patches on a series of object detectors individually and evaluate their ability to generalize across different models. In addition, we present the results of an ensemble model trained on all detectors. The results in Table \ref{tab:cross-model} show that our generated patches have a higher attack performance than Hu \etal \cite{NPAP}. Compared to Thys \etal \cite{thys2019fooling}, despite only being comparable to the attack performances of their method, our patches' naturalness surpasses theirs.

\subsection{Attack various detectors}
\label{sec:various-detector}
To better compare with the previous approach, we train and test our patch using the INRIA dataset on the same detectors and achieve the results in Table \ref{tab:inria}. Our method yields the strongest performance on all detectors. The qualitative user study in Sec. \ref{sec:user-study} further shows a large portion of users prefer our generated patches.

\subsection{User study}
\label{sec:user-study}
We conduct a user study to gather subjective scores of image patches to provide concrete evidence of the superiority of our approach. The patches used in the study are obtained from our method, previous methods \cite{NPAP,UPC,legitimate,thys2019fooling,makingcloak,adversarial_tshirt}, and real-world images. The results are obtained from 114 participants with diverse backgrounds. The screenshots of the survey interface can be found in Appendix.

In the first half of the questionnaire, the participants are shown several examples that demonstrate our definition of naturalistic patch. They are then asked to rate a series of image patches as natural or unnatural. The score for each test image is calculated by the ratio of positive votes. The results can be viewed in Table \ref{tab:naturalness_test}. The performance of our patch is the closest to the one of the real-world image, demonstrating its superiority over existing methods. 

In the second half of the survey, we perform a dedicated comparison between our method and a closely related work from Hu \etal \cite{NPAP}. For each question, we show an image patch from both methods and ask the participants to decide which one is more natural. Both images in each question are trained on the same object detector. Combined with Sec. \ref{sec:various-detector}, the results in Table \ref{tab:inria} show that our patches are generally more favorable among the participants and maintain superior attack capability. More subjective evaluation results can be found in Appendix.

\subsection{Cross-dataset evaluation}
The effectiveness of the generated adversarial patches on different datasets is evaluated using the MPII dataset. The results in Table \ref{tab:cross-dataset} indicate that the performance of the generated patches on MPII may not necessarily generalize well to other datasets, possibly because there are differences in attributes such as the position, number, viewing angle of people, \etc.

\subsection{Attack in physical world}
To evaluate the effectiveness of the generated adversarial patches in real-world scenes, we print a 40cm $\times$ 30cm patch on T-shirts. We use Logitech V-U0015 Webcam Camera (720p/30fps) to film two subjects with or without the ``adversarial T-shirt'' in various environments, including indoor and outdoor settings. During filming, we ask the subjects to move back and forth and side to side, while standing approximately two meters away from the camera. The mAP is calculated only on the subjects who wear adversarial T-shirt in all frames, so we can compare our method with previous works in the identical setting. The results shown in Table \ref{tab:physical_attack} demonstrates that our adversarial T-shirt significantly reduces the mAP to 23.33\% and 22.72\% in indoor and outdoor scenes, respectively. Compared to Hu \etal \cite{NPAP}, our adversarial T-shirt concretely beats theirs in both scenes.

\subsection{Robustness to existing defenses}
Liu \etal \cite{SAC} proposed Segment and Complete (SAC) for defending object detectors against patch attacks, allowing the object detector to bypass the adversarial patch and correctly detect the object. We conduct experiments to evaluate the effectiveness of the SAC in defending against our naturalistic adversarial patches. Our results, as presented in Table \ref{tab:sac}, suggest that our patches are natural enough to fail SAC. This is possibly due to the fact that their model was trained on adversarial noise patches, which may limit its ability to generalize on other types of adversarial patches.

\begin{table}
\centering
\caption{\label{tab:sac} Attack performance (mAP\%) against SAC. Best results are in \textbf{bold}.}
\begin{tabular*}{\linewidth}{@{\extracolsep{\fill}} lcc}
\toprule
Method & Vanilla YL2 & YL2 + SAC \cite{SAC} \\
\midrule
Thys \etal \cite{thys2019fooling} &2.68&24.53 ($\uparrow$ 21.85) \\
Hu \etal \cite{NPAP} &12.06&16.32 ($\uparrow$ 4.26) \\
Ours &11.62&\textbf{11.62 ($\uparrow$ 0.00)} \\
\bottomrule
\end{tabular*}
\end{table}

\section{Ablation study}
In Sec. \ref{sec:maintain-natural}, we introduced several components to promote naturalness in the generated patches. We thoroughly investigate the effect of these components on the attack performance and the naturalness preserving in this section.


\begin{table}
\centering
\caption{\label{tab:different-prompt} Attack performance (mAP\%) with different conditions.}
\begin{tabular*}{\linewidth}{@{\extracolsep{\fill}} ccccc}
    \toprule
    YL2 & YL4 & \multicolumn{3}{c}{YL4t} \\
    \cmidrule(lr){1-1}\cmidrule(lr){2-2}\cmidrule(lr){3-5}
    Cat& Princess& Car &Dog& \makecell{Uncond.\\after init.}\\
    \includegraphics[width=0.195\linewidth]{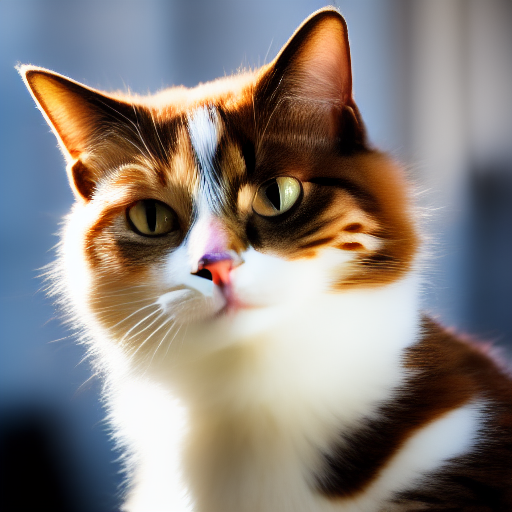}
    & \includegraphics[width=0.195\linewidth]{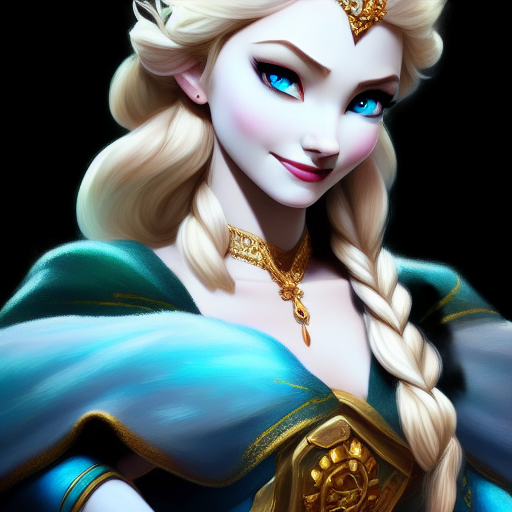}
    & \includegraphics[width=0.195\linewidth]{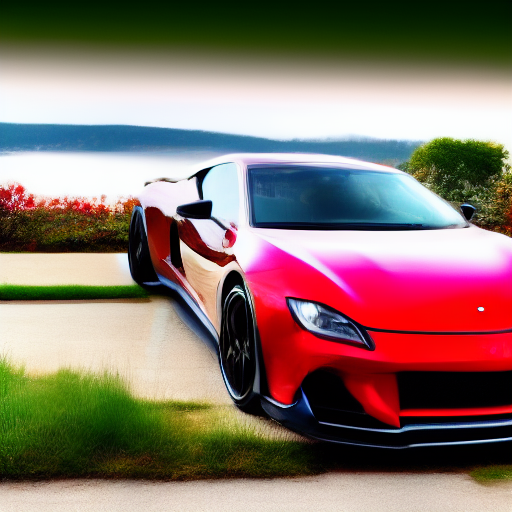}
    & \includegraphics[width=0.195\linewidth]{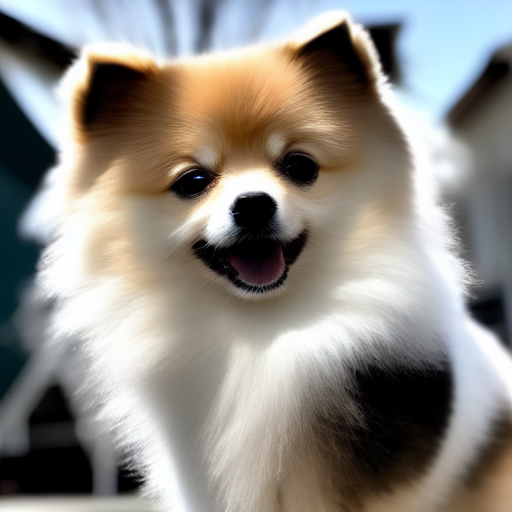}
    & \includegraphics[width=0.195\linewidth]{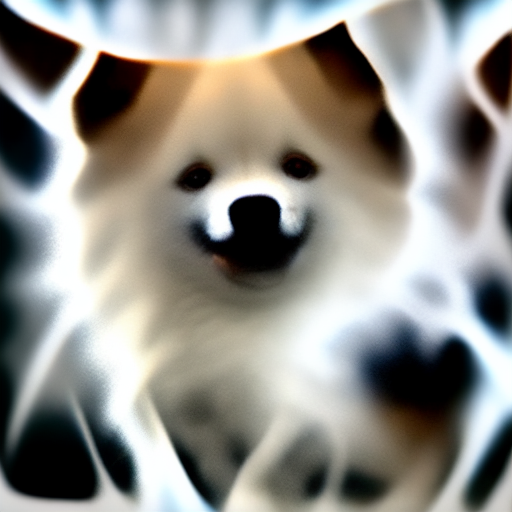} \\
    16.89& 24.47& 7.75 & 8.45 & 8.21\\
\bottomrule
\end{tabular*}
\end{table}

\subsection{Diffusion parameters}
\paragraph{Noise level.}
Through our experiments, we find that different noise levels significantly affect the training results.  We observe that setting $t_{\texttt{start}}$ to a small value (typically less than $0.3T$) does not introduce sufficient randomness to the process, resulting in images similar to those in prior study \cite{NPAP} which generates patches by adjusting the latent directly. 
Conversely, setting $t_{\texttt{start}}$ to a large value (usually greater than $0.7T$) leads to excessive randomness, causing gradient vanishing and prolonging training time. These effects are particularly pronounced at extreme values of $t$ ($0$ and $T$), where setting $t_{\texttt{start}}=0$ eliminates noise altogether, and setting $t_{\texttt{start}}=T$ results in pure noise and gradient elimination.


Therefore, in our experiments, we set $t \in [0.4T, 0.6T]$ as the ``sweet spot'', since it strikes a balance between introducing enough randomness and avoiding excessive noise that results in optimization difficulties.

\paragraph{Step size.}
As mentioned in Sec. \ref{sec:ddim}, the sampling method in Song \etal \cite{Song_ddim} allows us to set an arbitrary number of the step size $s$ in the reverse process, thus saving us time and memory while maintaining similar image quality. Based on experimental observations in \cite{Song_ddim} and ours, the image quality saturates as $s$ decreases. The improvement of quality cannot compensate the heavy computation loads using a low $s$ in our method.

\begin{figure}
    \begin{tikzpicture}[scale=0.88]
        \foreach \n/\i in {2/0,3/1,4/2,5/3,6/4,7/5,8/6}{
            \node at (\i,3.8) {\small 0.\n};
            \foreach \u/\j in {3/3,5/2,10/1,20/0}{
                \node at (\i,\j) {\includegraphics[width=0.1\linewidth]{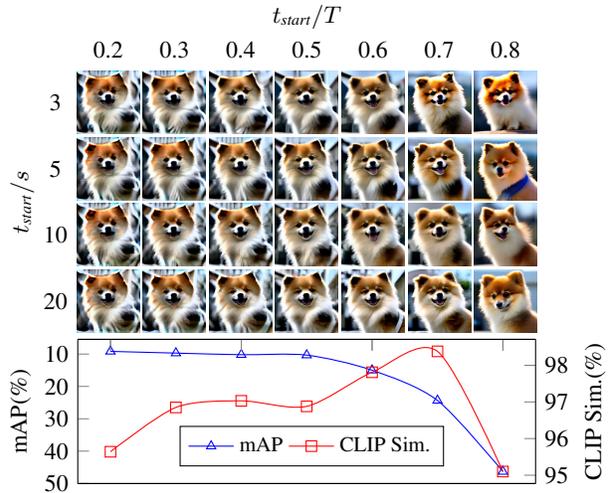}};
            }
        }
        \foreach \u/\j in {3/3,5/2,10/1,20/0} {
            \node at (-0.8,\j) {\small \u};
        }
        \node at (3,4.3) {\small $t_{\textit{start}}/T$};
        \node at (-1.3,1.5) {\rotatebox{90}{\small $t_{\textit{start}}/s$}};
    \end{tikzpicture}
    \vskip -0.3cm
    \begin{tikzpicture}
        \pgfplotsset{
            every tick label/.append style={font=\small},
            yticklabel={\pgfmathprintnumber[assume math mode=true]{\tick}}
        }
        \begin{axis}[
            xmin=0.15, xmax=0.85,
            xtick pos=top,
            axis y line*=left,
            xticklabels={,,},
            width=0.92\linewidth,
            height=3.5cm,
            ylabel={\small mAP(\%)},
            y label style={at={(axis description cs:0.12,.5)},anchor=south},
            y dir=reverse
        ]
            \addplot[smooth, mark=triangle, blue] table [x=t_start, y=mAP, col sep=comma] {figs/noise-n-step/stats.csv};
            \label{plot:mAP}
        \end{axis}
        
        \begin{axis}[
            xmin=0.15, xmax=0.85,
            xtick pos=top,
            axis y line*=right,
            xticklabels={,,},
            width=0.92\linewidth,
            height=3.5cm,
            ylabel={\small CLIP Sim.(\%)},
            legend columns=-1,
            y label style={at={(axis description cs:1.375,.5)},anchor=south},
            legend style={at={(0.8,0.4)}}
        ]
            \addlegendimage{/pgfplots/refstyle=plot:mAP}\addlegendentry{\footnotesize mAP}
            \addplot[smooth, mark=square, red] table [x=t_start, y=clip_sim, col sep=comma] {figs/noise-n-step/stats.csv};
            \addlegendentry{\footnotesize CLIP Sim.}
        \end{axis}
    \end{tikzpicture}
    \caption{Effects of the noise level and step size. In the bottom line chart, each data point is an average score of the four patches on the same vertical axis.}
    \label{fig:noise-n-step}
\end{figure}

We provide a holistic view for the effect of $t_{\texttt{start}}$ and $s$ in Figure \ref{fig:noise-n-step}, where we use CLIP \cite{Clip} embeddings to calculate the similarity between the initial patch $P_{\textit{init}}$ and the optimized patch. CLIP is a foundation model frequently used as the go-to feature extractor. We use the similarity as an automatic objective metric that reflects the preservation of image quality and semantic. We visually show that $s$ has little impact on the generated patch and therefore average the statistics along its axis in the line chart of Figure \ref{fig:noise-n-step}. On the contrary, small $t_{\textit{start}}$ leads to suboptimal image quality due to its lack of randomness, and big $t_{\textit{start}}$ causes the image to disregard our optimization objective and its semantic, despite generating good-looking pictures.

\vspace{0.1cm}
\subsection{Text condition}
As we have claimed in our method, text condition $c_P$ plays an important role in the maintenance of image naturalness. We show that our framework provides a great degree of freedom in choosing the desired prompt and subject for the patch. As long as $c_P$ guides the semantic of the image during optimization, we generally obtain a naturalistic patch that can be used to attack detectors. Conversely, without conditioning on the text prompt during gradient updates, the patch loses its naturalness in order to reach comparable attack capability, as shown in Table \ref{tab:different-prompt}.

\vspace{0.1cm}
\section{Conclusion}
In this work, we develop the first naturalistic adversarial patch generation method based on DM. We demonstrate the DM pretrained upon large-scale natural images, \ie Stable Diffusion \cite{LDM}, is significantly useful to generate high-quality and naturalistic adversarial patches while achieving promising attack performance. In this paper, we further present many practical experiences and trade-offs using DM to generate an effective adversarial patch under different conditions. The evaluation results of extensive quantitative, qualitative, and subjective experiments have demonstrated the effectiveness of the proposed approach as compared with other state-of-the-art patch generation methods.







{\small
\bibliographystyle{ieee_fullname}
\bibliography{egbib}
}

\pagebreak


\appendix
\section{Experiment details}
In this section, we provide additional experiment details for generating the proposed adversarial patches.
\paragraph{Optimization hyperparameters.}
In our main objective function of the adversarial patch training,

\begin{equation}
    \min_{P} \left( \frac{1}{N} \sum^N_{i=1} \mathcal{L}_{det}(\mathcal{I}^i_{P'})+\lambda\cdot \mathcal{L}_{tv}(P') \right),
\end{equation}
we set $\lambda$ to 0.1 to control the effect of total variation loss. During training, we use a batch size $N$ of 8 for most detectors, with the exception of YOLOv3 and YOLOv4, for which we use $N$ of 4 and 1, respectively, due to their higher memory requirements.

\paragraph{Datasets.}
We adopt the datasets used by Hu \etal \cite{NPAP}, including INRIA \cite{INRIA} and MPII \cite{MPII} datasets. We follow the same protocol in \cite{NPAP}, in which we use the object detectors' predictions on clean test data (i.e., without applying any patch) as the reference labels. Table \ref{tab:inria-mpii} shows examples of the test images in both datasets and their corresponding reference labels. 

\paragraph{Pretrained models.}
We employ pretrained weights for both the detectors and the diffusion model, which are kept fixed during training. To encourage reproducibility we provide the details of the object detectors and diffusion models utilized in our experiments in Table \ref{tab:model-link}.



\paragraph{Text conditioning}
We use text conditioning to guide the patch generation, which helps maintaining the naturalness of the patch during the search for adversaries. In Table \ref{tab:prompt}, we provide selected examples of detailed text prompts that we use to generate the initial patches.


\section{Physical experiments}
\begin{figure*}
        \centering
        \includegraphics[width=5.5cm]{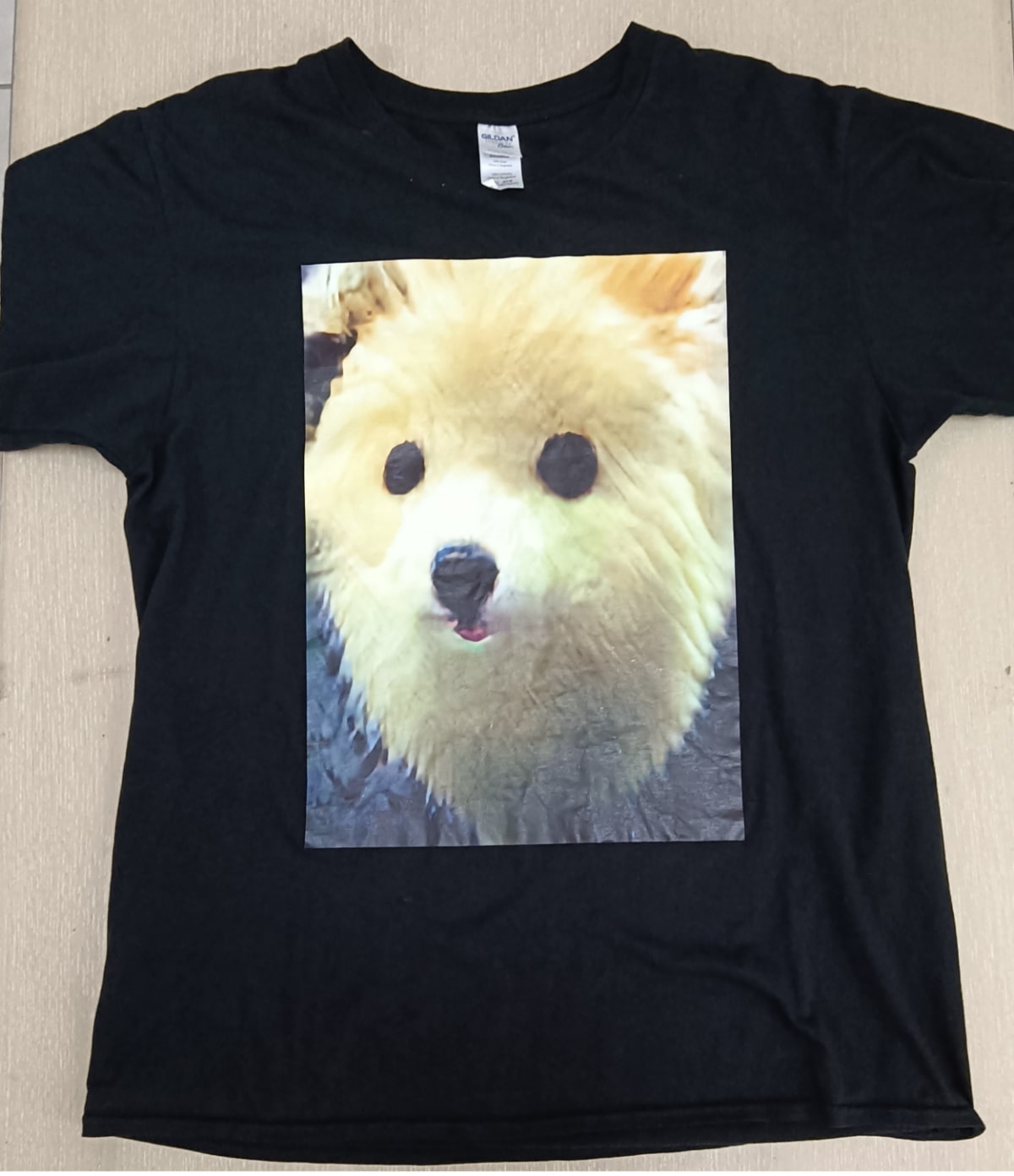}
        \includegraphics[width=5.5cm]{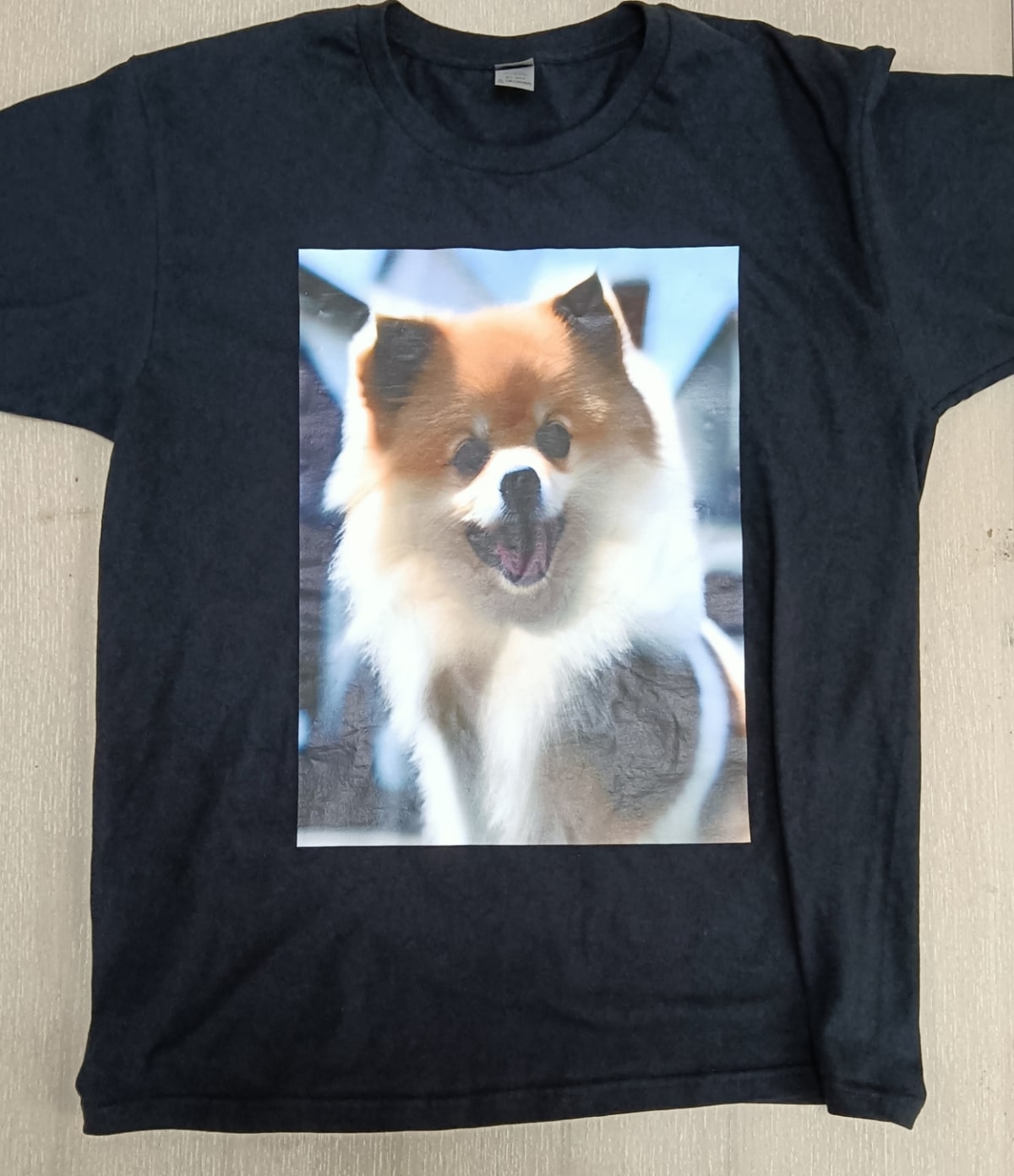}
        \includegraphics[width=5.5cm]{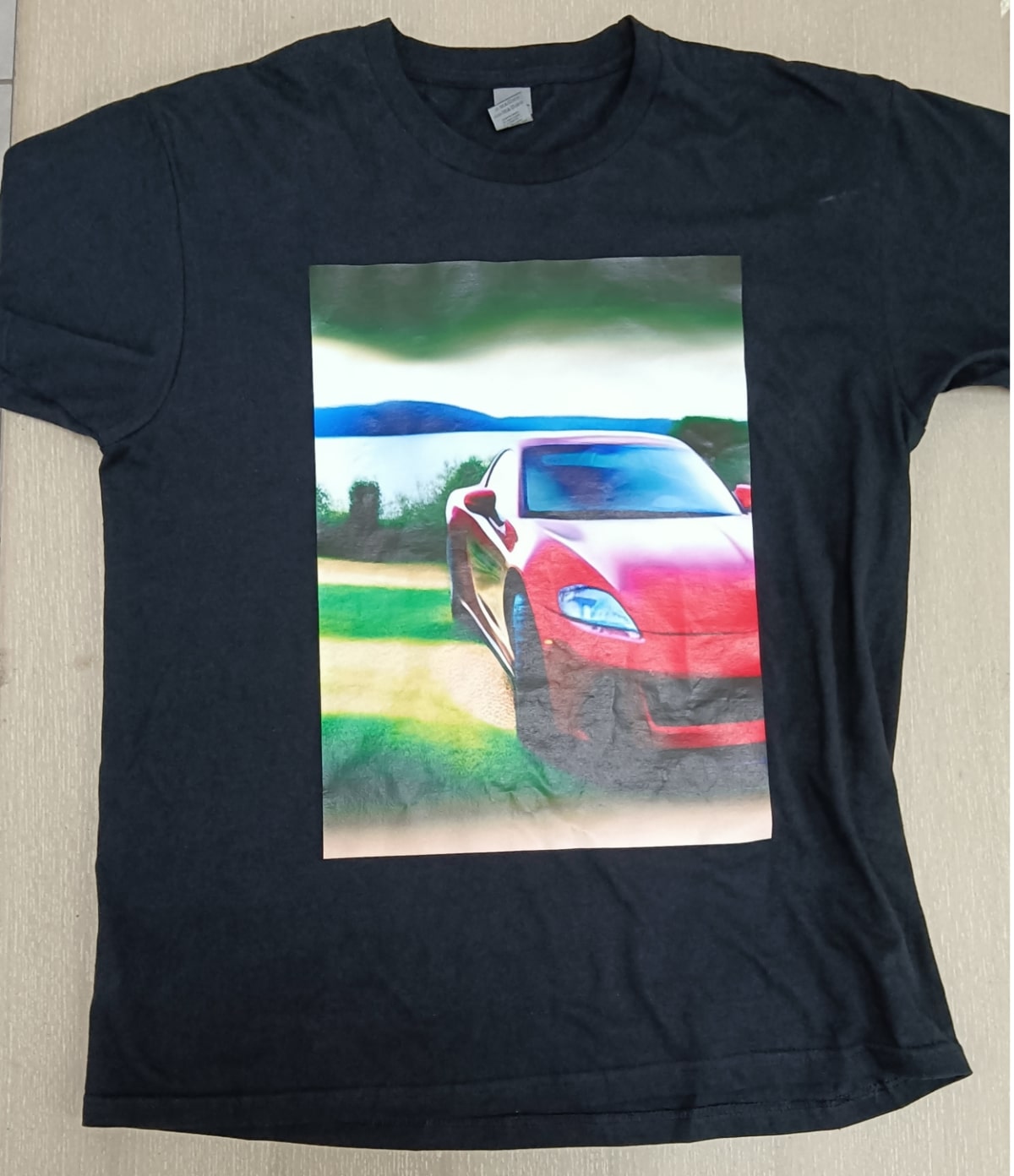}\\
        \includegraphics[width=5.5cm]{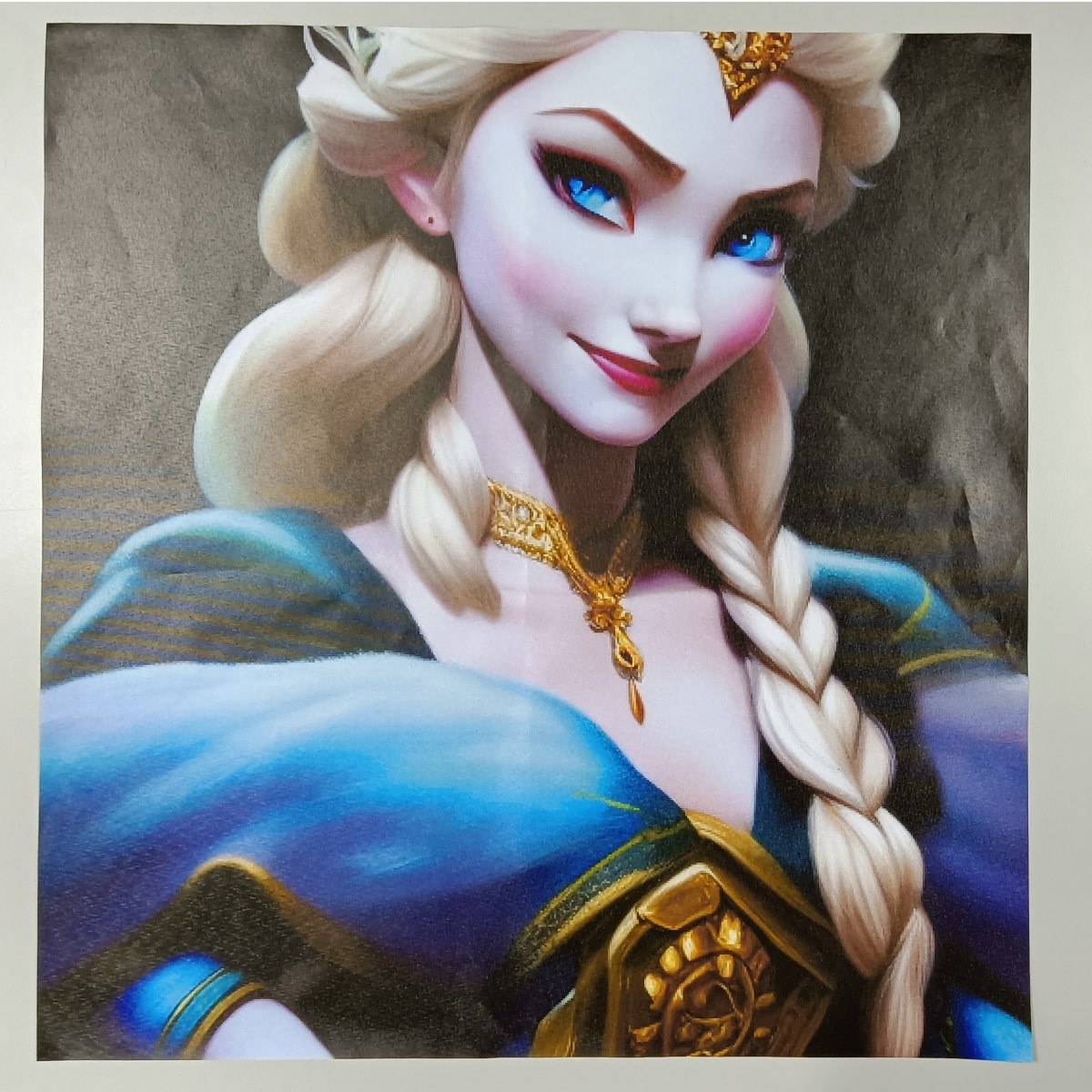}
        \includegraphics[width=5.5cm]{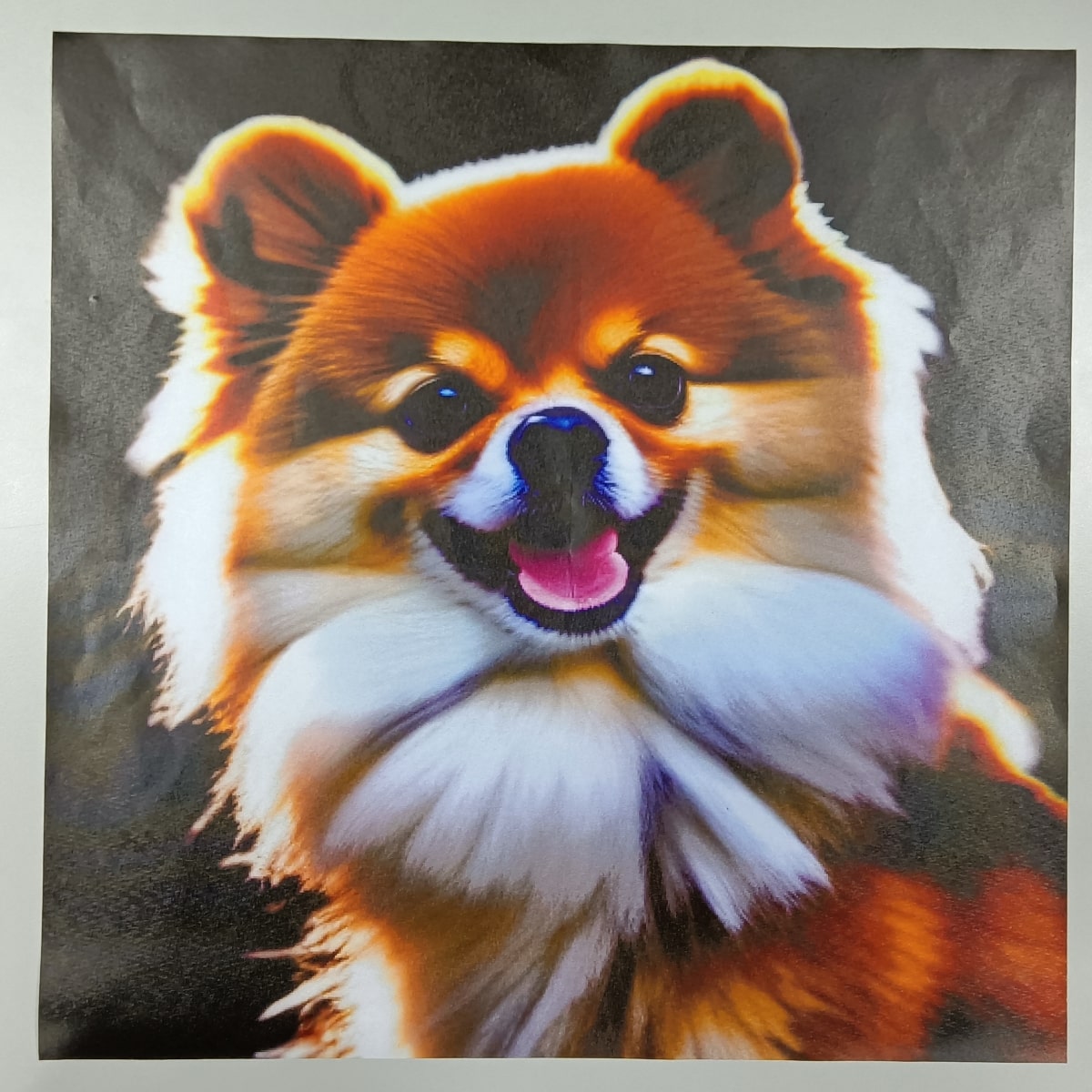}
        \includegraphics[width=5.5cm]{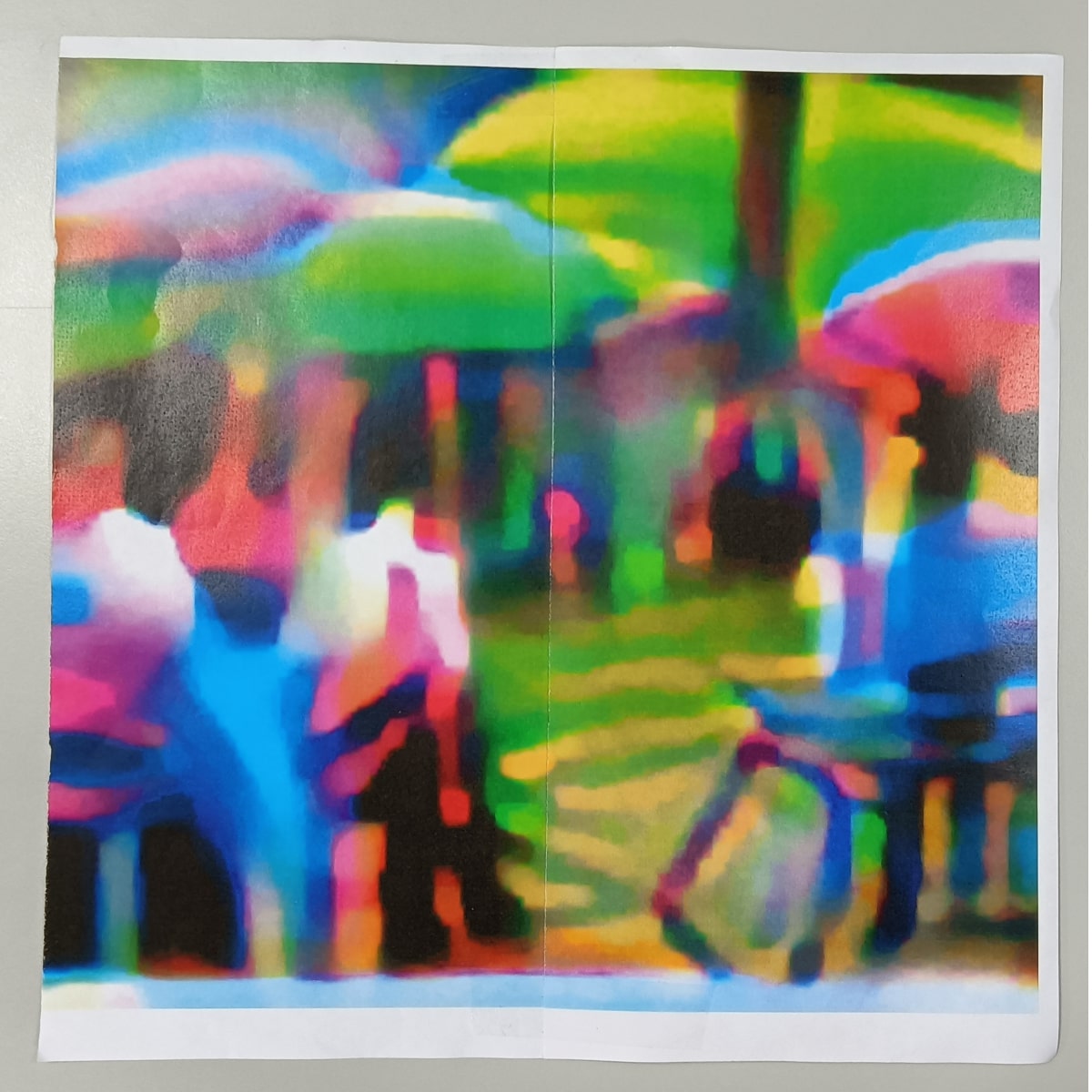}\\
        \caption{\textbf{Printed materials, from top left to right bottom are cloth, cloth, cloth, canvas, canvas, and paper, respectively.}}
        \label{fig:materials}
\end{figure*}

\begin{figure*}
        \centering
        \includegraphics[width=4.2cm]{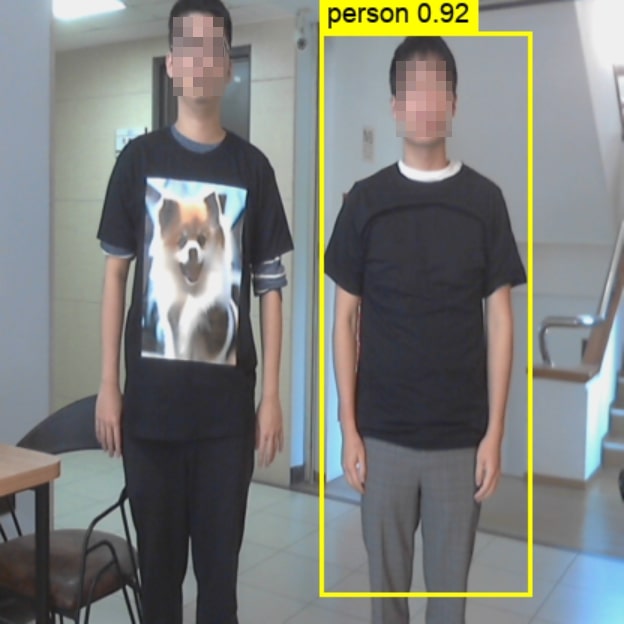}
        \includegraphics[width=4.2cm]{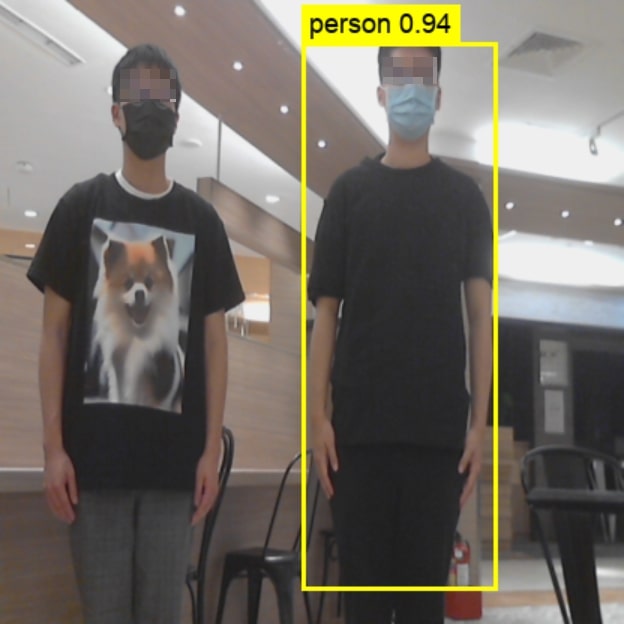}
        \includegraphics[width=4.2cm]{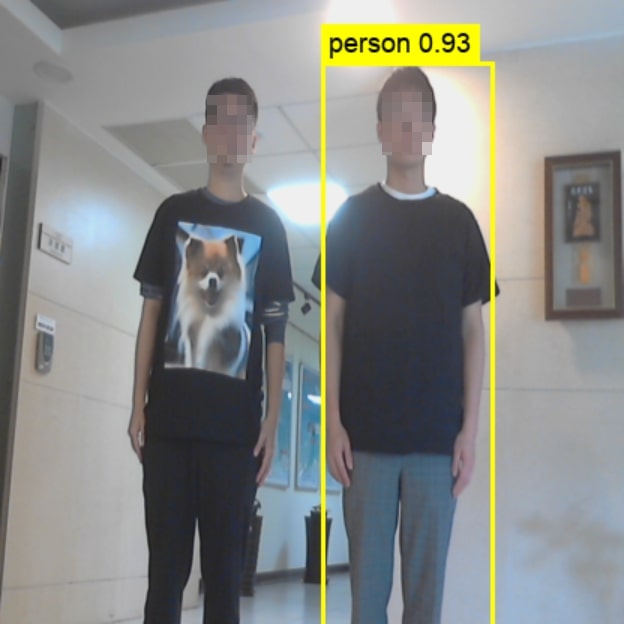}
        \includegraphics[width=4.2cm]{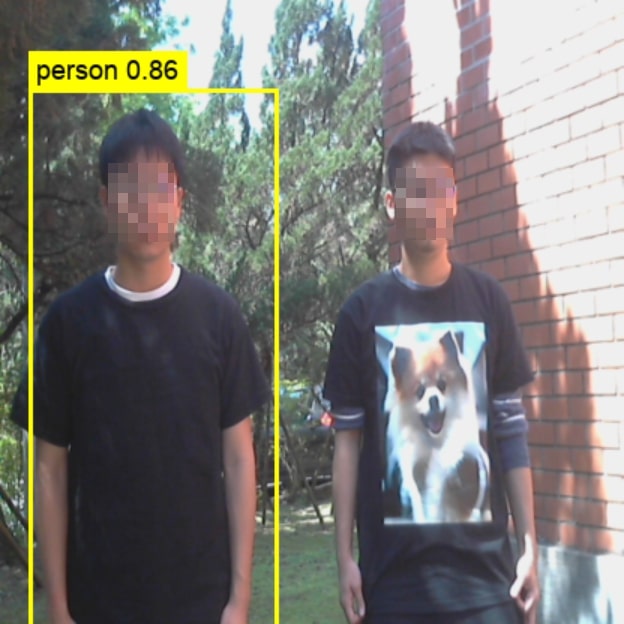}\\
        \includegraphics[width=4.2cm]{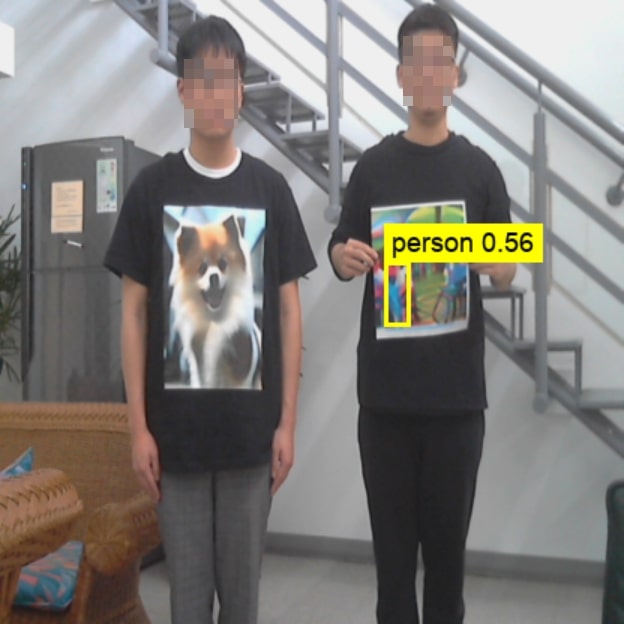}
        \includegraphics[width=4.2cm]{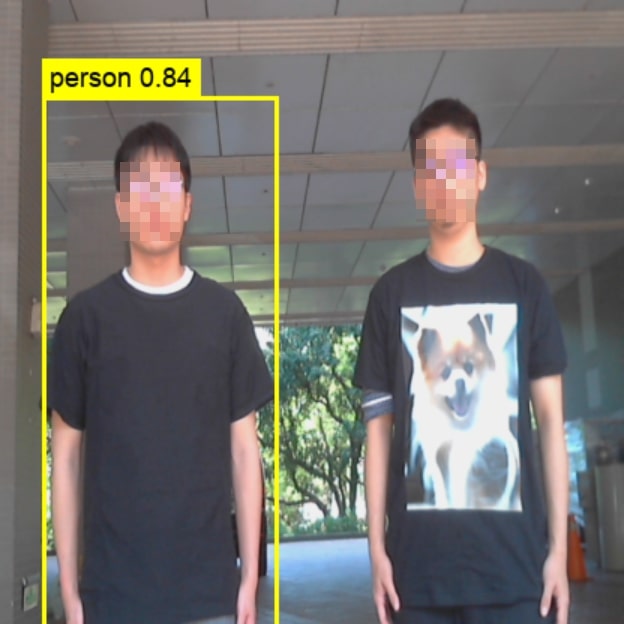}
        \includegraphics[width=4.2cm]{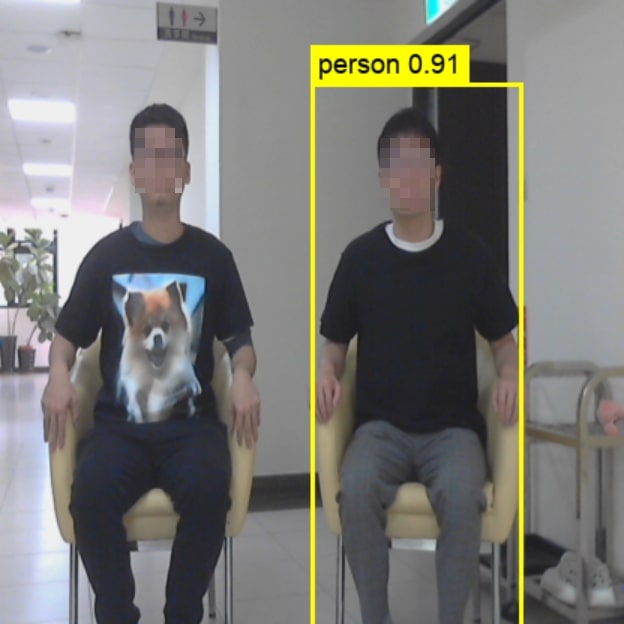}
        \includegraphics[width=4.2cm]{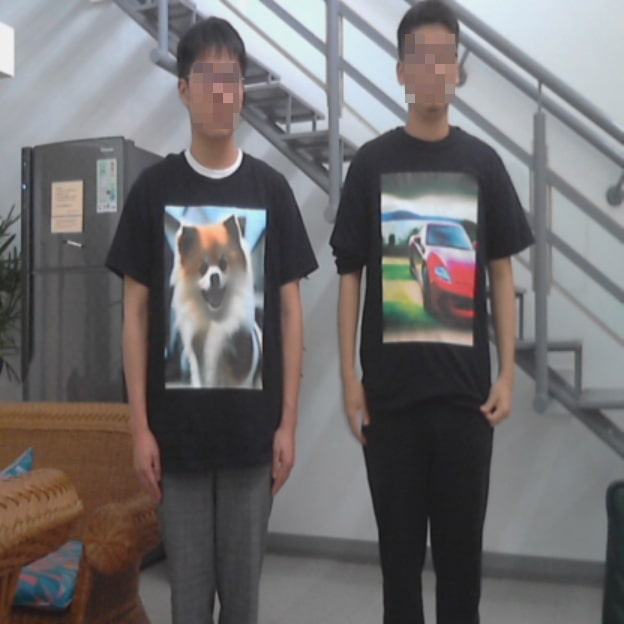}\\
        \includegraphics[width=4.2cm]{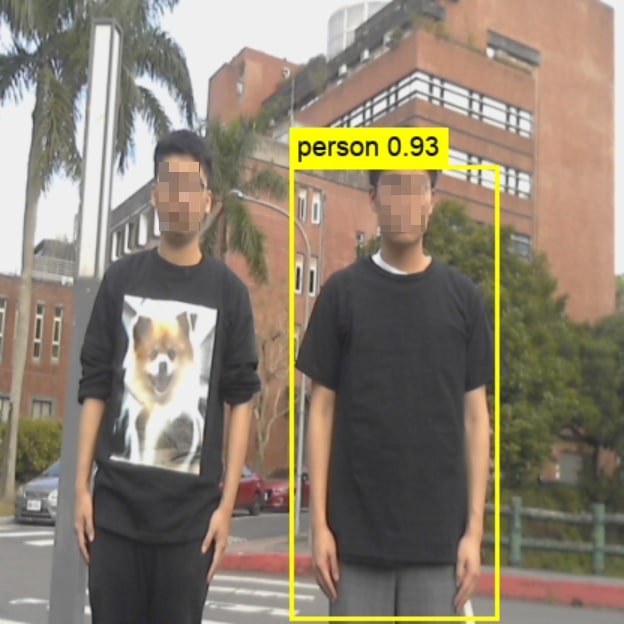}
        \includegraphics[width=4.2cm]{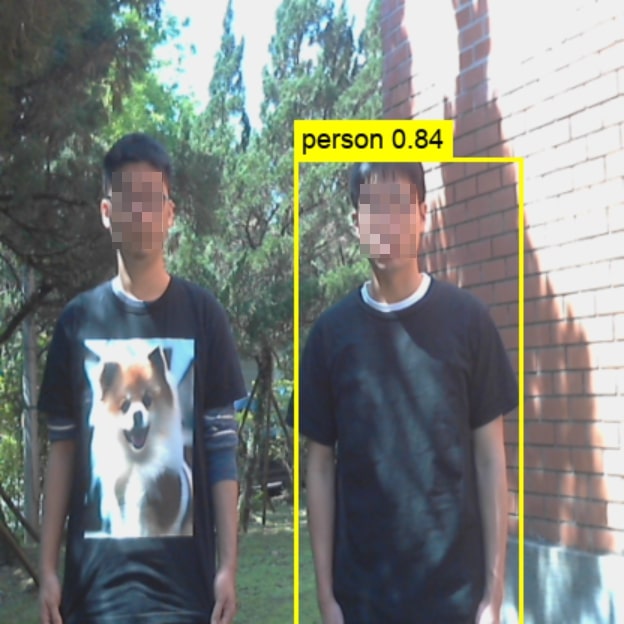}
        \includegraphics[width=4.2cm]{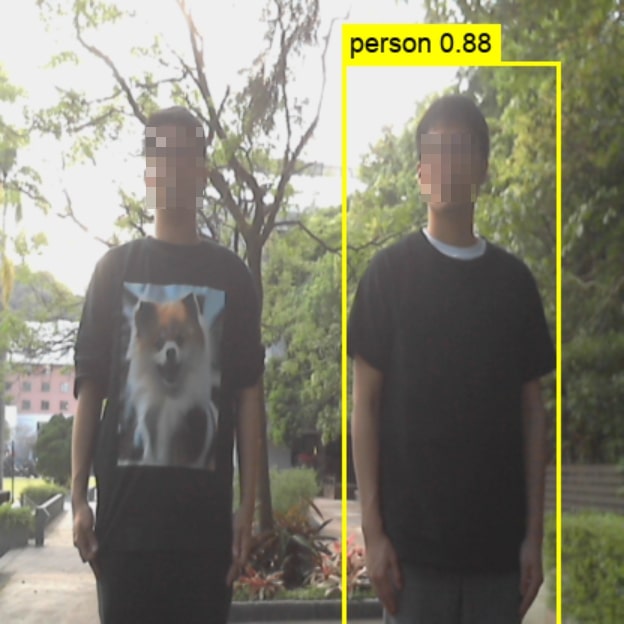}
        \includegraphics[width=4.2cm]{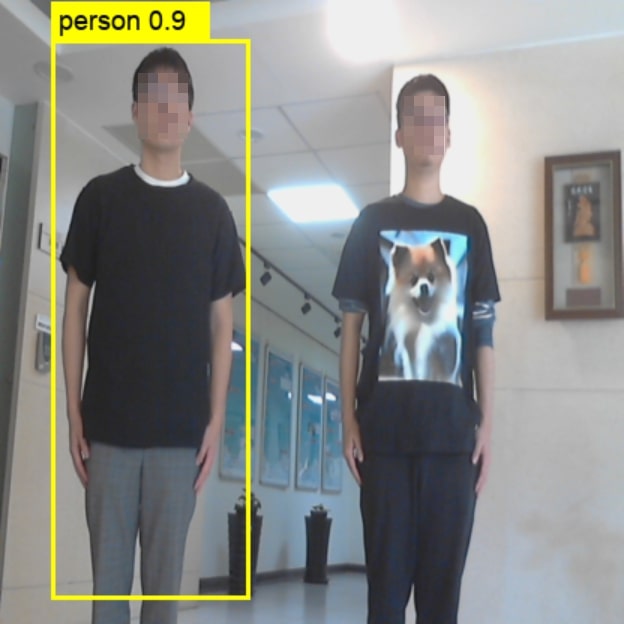}\\
        \includegraphics[width=4.2cm]{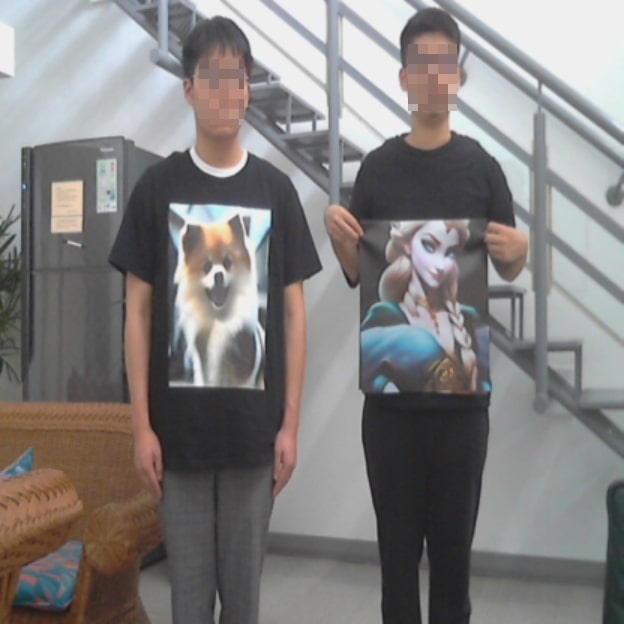}
        \includegraphics[width=4.2cm]{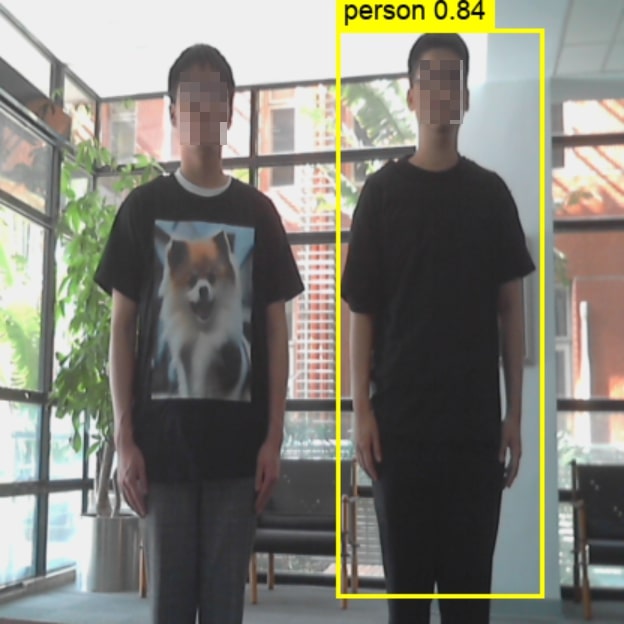}
        \includegraphics[width=4.2cm]{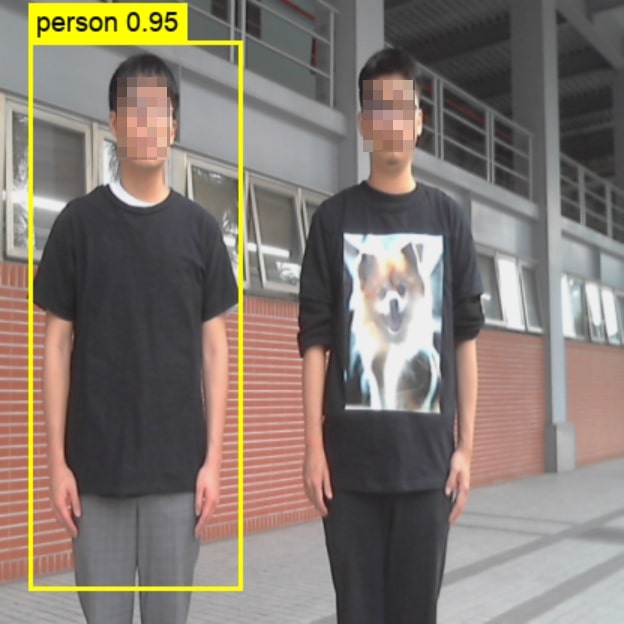}
        \includegraphics[width=4.2cm]{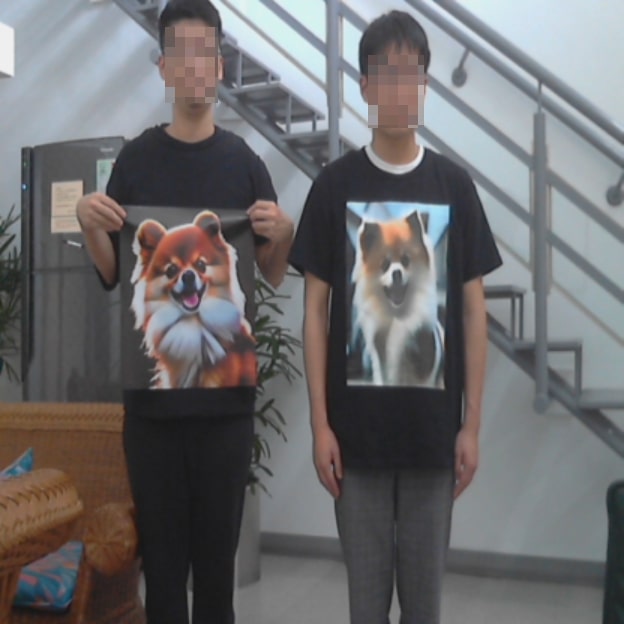}\\
        \includegraphics[width=4.2cm]{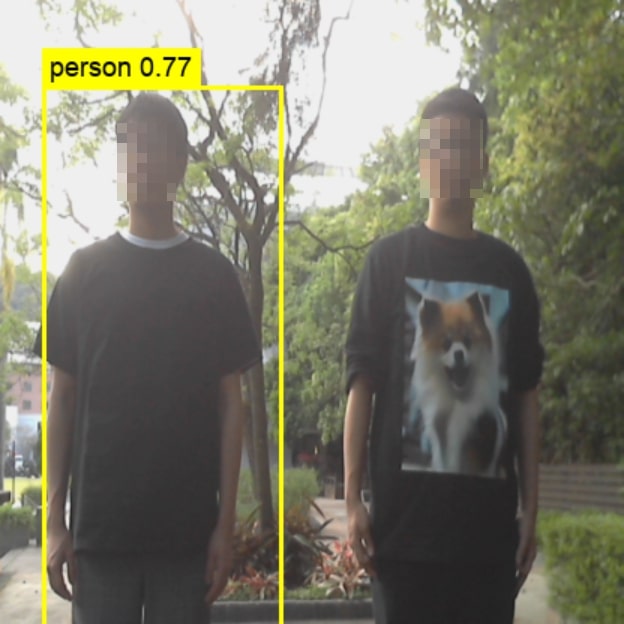}
        \includegraphics[width=4.2cm]{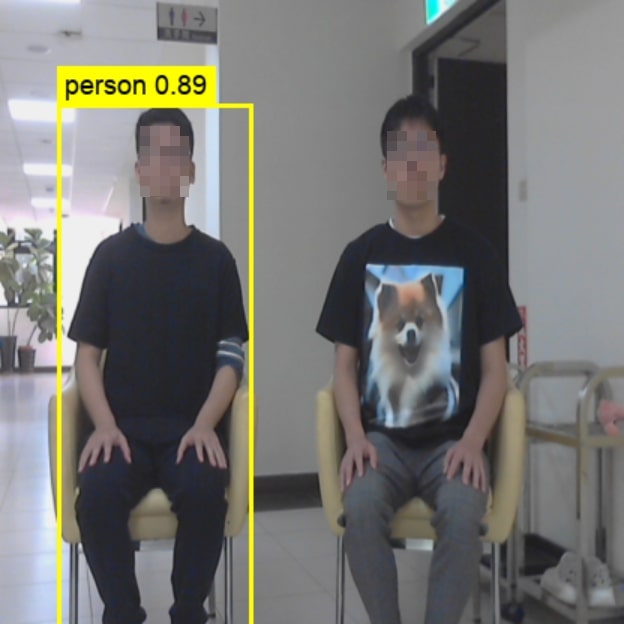}
        \includegraphics[width=4.2cm]{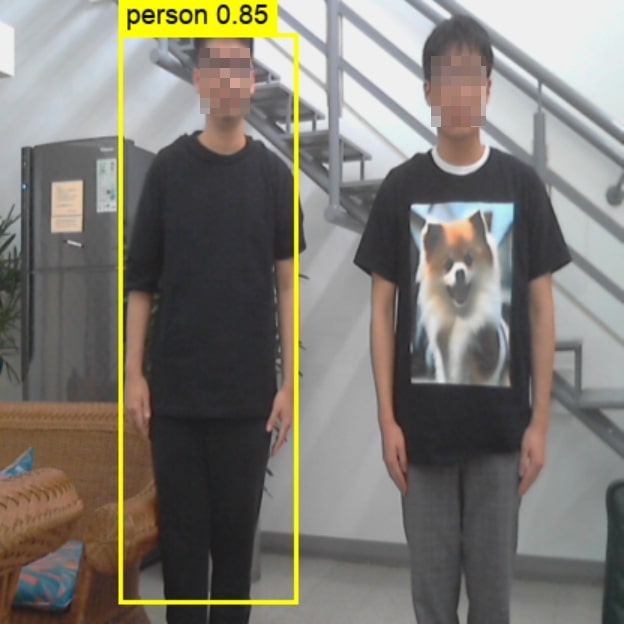}
        \includegraphics[width=4.2cm]{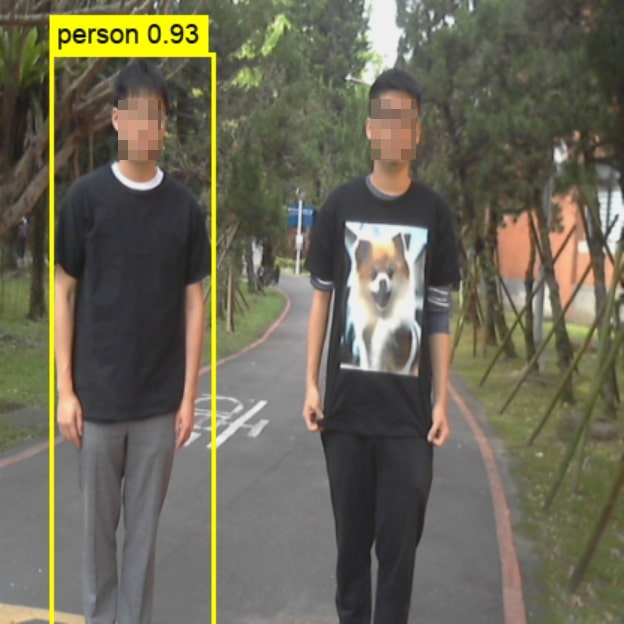}\\
        \caption{\textbf{Additional physical evaluation images under different scenes.}}
        \label{fig:more_physical}
\end{figure*}

To show that our method generalizes well to real-world settings, we use commercial heat transfer technology to print the adversarial patches onto T-shirts. In addition to clothes, we also realize the adversarial patch as canvas and paper. Figure \ref{fig:materials} shows pictures of the printed materials. We also provide additional images to further demonstrate the real-world scenes in our physical experiments in Figure \ref{fig:more_physical}.
\setlength\tabcolsep{0pt}
\begin{table}
\centering
\caption{\label{tab:inria-mpii} \textbf{Samples from the INRIA \cite{INRIA} and MPII datasets \cite{MPII}}.}
\begin{tabular*}{\linewidth}{@{\extracolsep{\fill}} ccc}
   \toprule
    \multicolumn{3}{c}{INRIA \cite{INRIA}}\\  \cmidrule(lr){1-3}
    \includegraphics[width=0.33\linewidth]{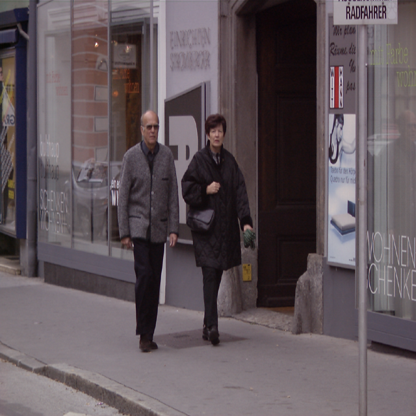}
    & \includegraphics[width=0.33\linewidth]{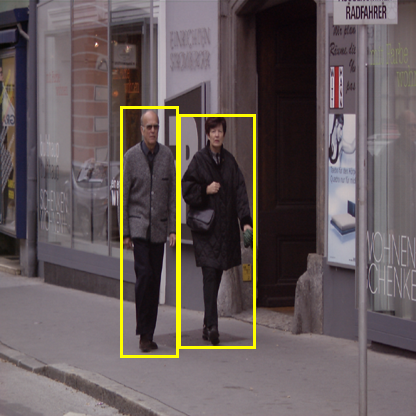}
    & \includegraphics[width=0.33\linewidth]{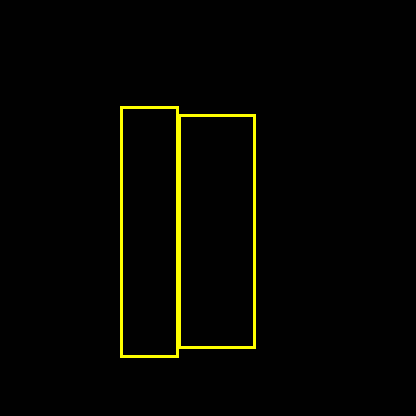}\\
    \midrule
    \multicolumn{3}{c}{MPII \cite{MPII}} \\ \cmidrule(lr){1-3}
    \includegraphics[width=0.33\linewidth]{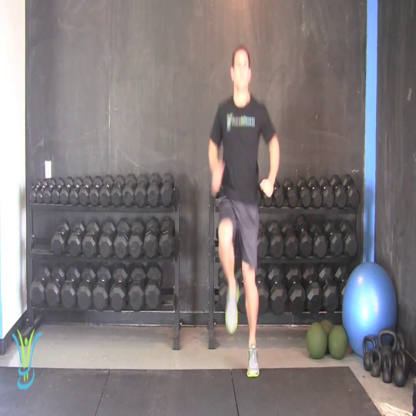}
    & \includegraphics[width=0.33\linewidth]{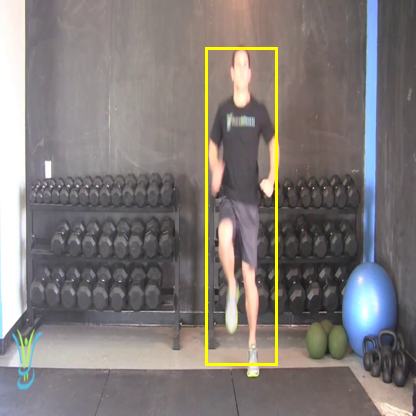}
    & \includegraphics[width=0.33\linewidth]{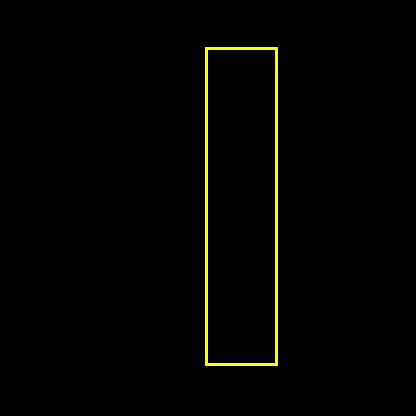} \\
\bottomrule
\end{tabular*}
\end{table}

\begin{table*}
\centering
\caption{\textbf{Model repositories and weights.}}
\label{tab:model-link}
\begin{tabular*}{\linewidth}{@{\extracolsep{\fill}} l p{13.5cm}}
\toprule
\multirow{2}{*}{Model family} & Code repository \\
\cmidrule(lr){2-2}
& Pretrained weights \\
\midrule
\midrule
\multirow{2}{*}{YOLOv2 \cite{yolov2}} & \url{https://gitlab.com/EAVISE/adversarial-yolo} \\
\cmidrule(lr){2-2}
& \url{https://pjreddie.com/media/files/yolov2.weights} \\
\midrule
\multirow{3}{*}{YOLOv3 \cite{yolov3}} & \url{https://github.com/eriklindernoren/PyTorch-YOLOv3} \\
\cmidrule(lr){2-2}
& \url{https://pjreddie.com/media/files/yolov3.weights} \\
& \url{https://pjreddie.com/media/files/yolov3-tiny.weights} \\
\midrule
\multirow{3}{*}{YOLOv4 \cite{yolov4}} & \url{https://github.com/Tianxiaomo/pytorch-YOLOv4} \\
\cmidrule(lr){2-2}
& \url{https://www.dropbox.com/s/jp30sq9k21op55j/yolov4.weights} \\
& \url{https://www.dropbox.com/s/t90a1xazhbh2ere/yolov4-tiny.weights} \\
\midrule
\multirow{3}{*}{YOLOv5 \cite{yolov5}} & \url{https://github.com/ultralytics/yolov5} \\
\cmidrule(lr){2-2}
& \url{https://github.com/ultralytics/yolov5/releases/download/v7.0/yolov5s.pt} \\
\midrule
\multirow{3}{*}{YOLOv7 \cite{yolov7}} & \url{https://github.com/WongKinYiu/yolov7} \\
\cmidrule(lr){2-2}
& \url{https://github.com/WongKinYiu/yolov7/releases/download/v0.1/yolov7-tiny.pt} \\
\midrule
\multirow{4}{*}{FasterRCNN \cite{fasterRCNN}} & \url{https://pytorch.org/vision/0.12/_modules/torchvision/models/detection/faster_rcnn.html} \\
\cmidrule(lr){2-2}
& \url{https://download.pytorch.org/models/fasterrcnn_resnet50_fpn_coco-258fb6c6.pth} \\
\midrule
\multirow{2}{*}{Deformable DETR \cite{D_Detr}} & \url{https://huggingface.co/SenseTime/deformable-detr} \\
\cmidrule(lr){2-2}
& \url{https://huggingface.co/SenseTime/deformable-detr} \\
\midrule
\multirow{3}{*}{Latent Diffusion \cite{LDM}} & \url{https://github.com/CompVis/latent-diffusion} \\
\cmidrule(lr){2-2}
& \url{https://ommer-lab.com/files/latent-diffusion/nitro/txt2img-f8-large/model.ckpt} \\
\midrule
\multirow{3}{*}{Stable Diffusion \cite{LDM}} & \url{https://github.com/CompVis/stable-diffusion} \\
\cmidrule(lr){2-2}
& \url{https://huggingface.co/CompVis/stable-diffusion-v1-4} \\
& \url{https://huggingface.co/hakurei/waifu-diffusion-v1-3}\\
& \url{https://huggingface.co/stabilityai/stable-diffusion-2-base} \\
\bottomrule
\end{tabular*}
\end{table*}

\begin{table*}
\centering
\caption{\label{tab:prompt} \textbf{Initialized images and their corresponding text prompts.}
}
\begin{tabular*}{\linewidth}{@{\extracolsep{\fill}} l p{5.7cm} p{5.7cm}}
   \toprule
    Generated image & Prompt & Negative Prompt \\
    \midrule
    
    \multirow{1}{*}{\includegraphics[width=0.1\linewidth]{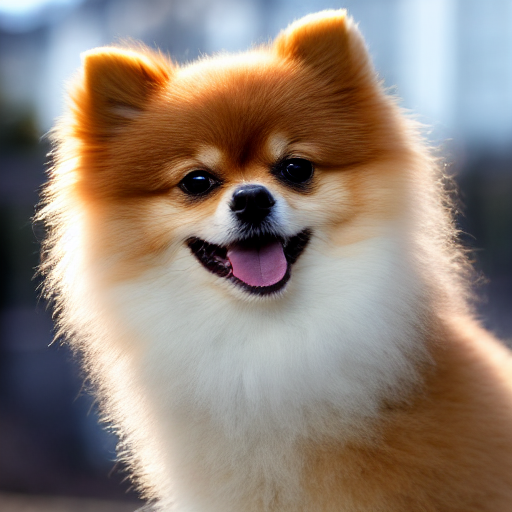}} &
    \texttt{A high quality photo of a Pomeranian, clean background.} &
    \texttt{Bad anatomy, bad proportions, deformed, disfigured, duplicate, mutation, ugly, weird eyes.} \\
    \cmidrule{2-3}
    \multirow{1}{*}{\includegraphics[width=0.1\linewidth]{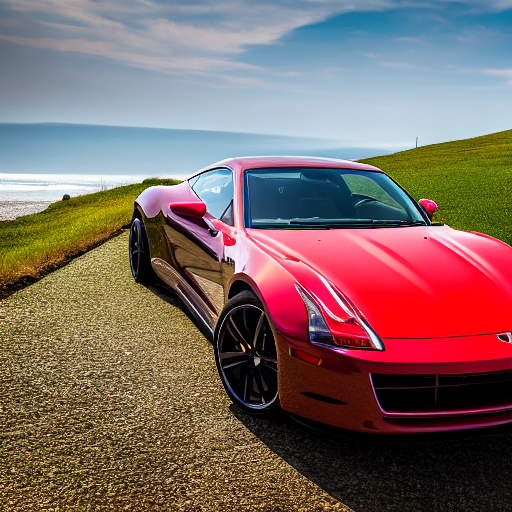}} &
    \texttt{A high quality photo of a car.} &
    \texttt{Bad anatomy, bad proportions, deformed, disfigured, duplicate, mutation, ugly.} \\
    \cmidrule{2-3}
    \multirow{1}{*}{\includegraphics[width=0.1\linewidth]{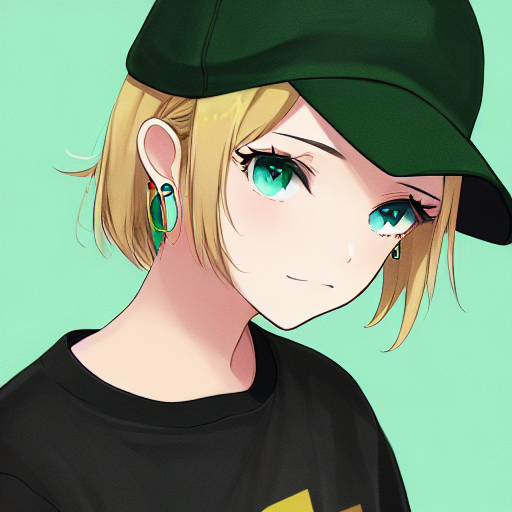}} &
    \texttt{1girl, aqua eyes, baseball cap, blonde hair, closed mouth, earrings, green background, hat, hoop earrings,jewelry, looking at viewer, shirt, short hair, simple background, solo, upper body, yellow shirt, without hand.} &
    \texttt{Bad anatomy, bad proportions, deformed, disfigured, duplicate, mutation, ugly, weird eyes.} \\
\bottomrule
\end{tabular*}
\end{table*}


\section{User study}
\begin{figure*}
        \centering
        \includegraphics[width=0.93\linewidth]{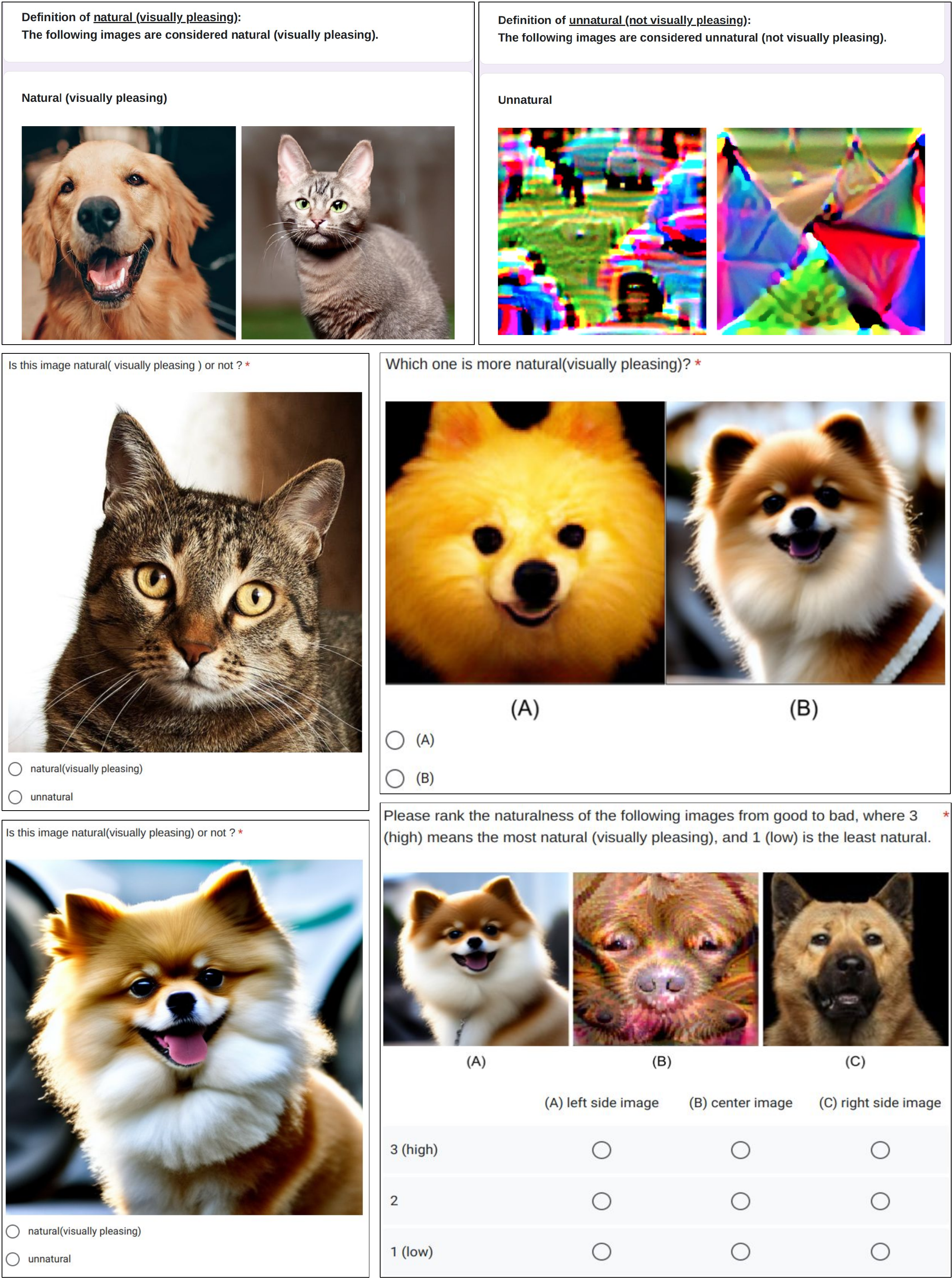}\\
        \caption{\textbf{Screenshots of the survey interface.}}
        \label{fig:screenshot}
\end{figure*}

We conduct a user study with 114 participants to support the superiority of our patches in terms of human preferences. Selected screenshots of the survey interface are shown in Figure \ref{fig:screenshot}. 

We have shown the patches from the first part of the user study in Table 3 of the main paper. In Table \ref{tab:vsNPAP}, we further show the patches used in the second part, where we compare our generated patches with the ones from Hu \etal \cite{NPAP}. As presented in Table 2 of the main paper, our attack method achieves better attack performance. Meanwhile, our adversarial patches are considered more natural in all surveys we conduct.

\begin{table*}
\centering
\caption{\label{tab:vsNPAP} \textbf{User preferences among patches in comparison to Hu \etal \cite{NPAP}} (Our patches hold either superior or similar mAP drops). The {\color[HTML]{696EC4} \textbf{purple}} color indicates the preference score of the left image (ours), and the {\color[HTML]{66C3FF} \textbf{blue}} indicates the right one (Hu \etal \cite{NPAP}).}
\begin{tabular*}{\linewidth}{@{\extracolsep{\fill}} ccc}
   \toprule

    \includegraphics[width=0.33\linewidth]{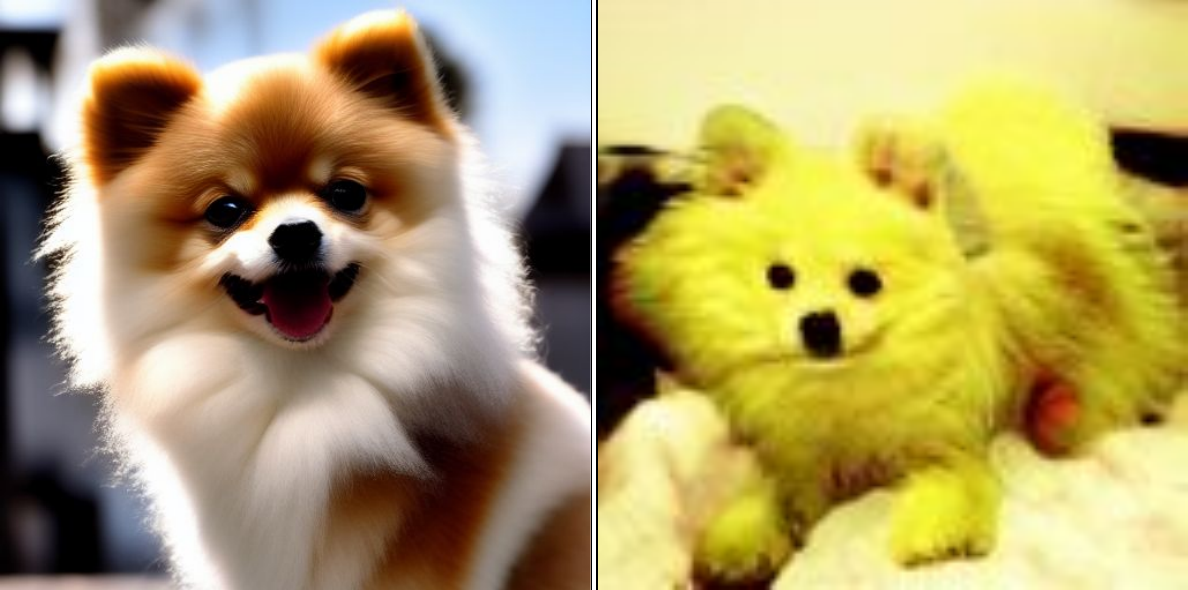}
    & \includegraphics[width=0.33\linewidth]{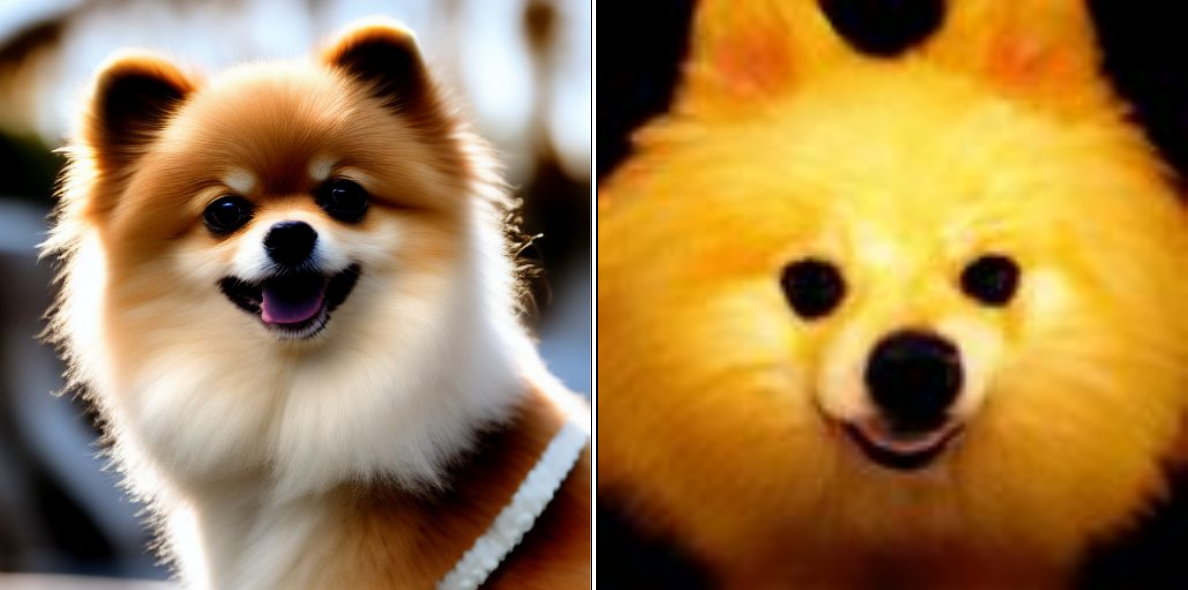}
    & \includegraphics[width=0.33\linewidth]{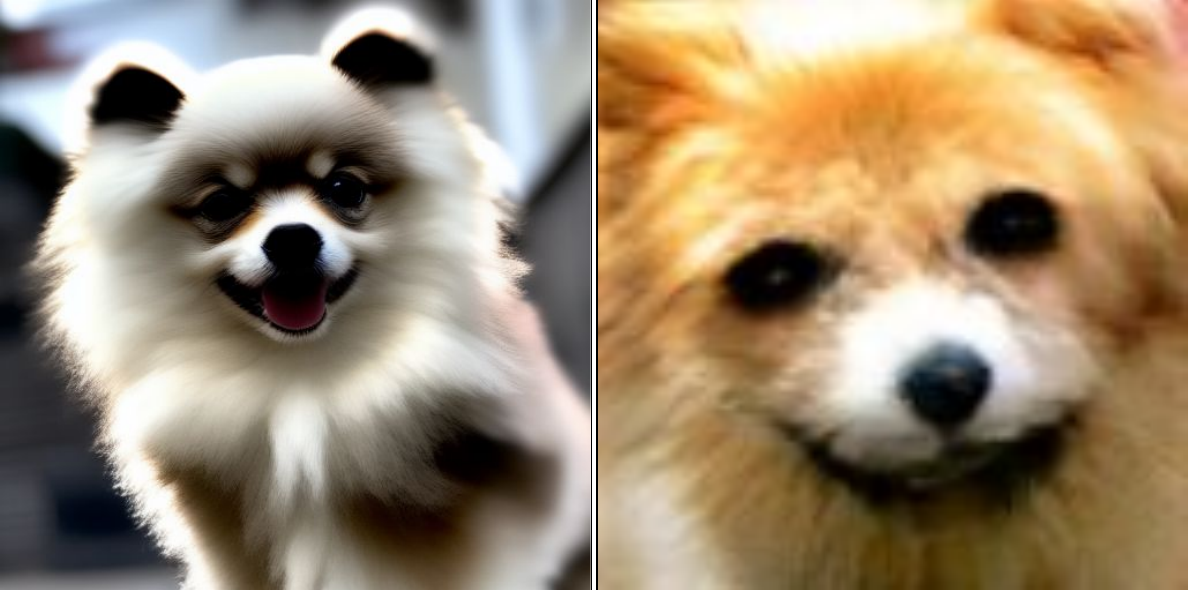}\\
    
    \includegraphics[width=0.33\linewidth]{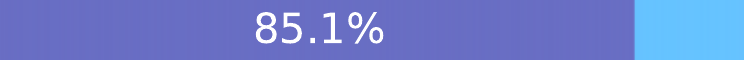}
    & \includegraphics[width=0.33\linewidth]{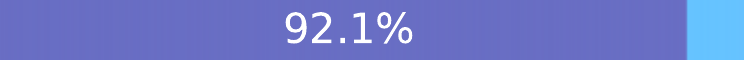}
    & \includegraphics[width=0.33\linewidth]{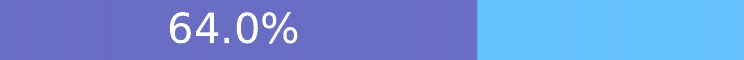}\\
    
    \midrule
    \includegraphics[width=0.33\linewidth]{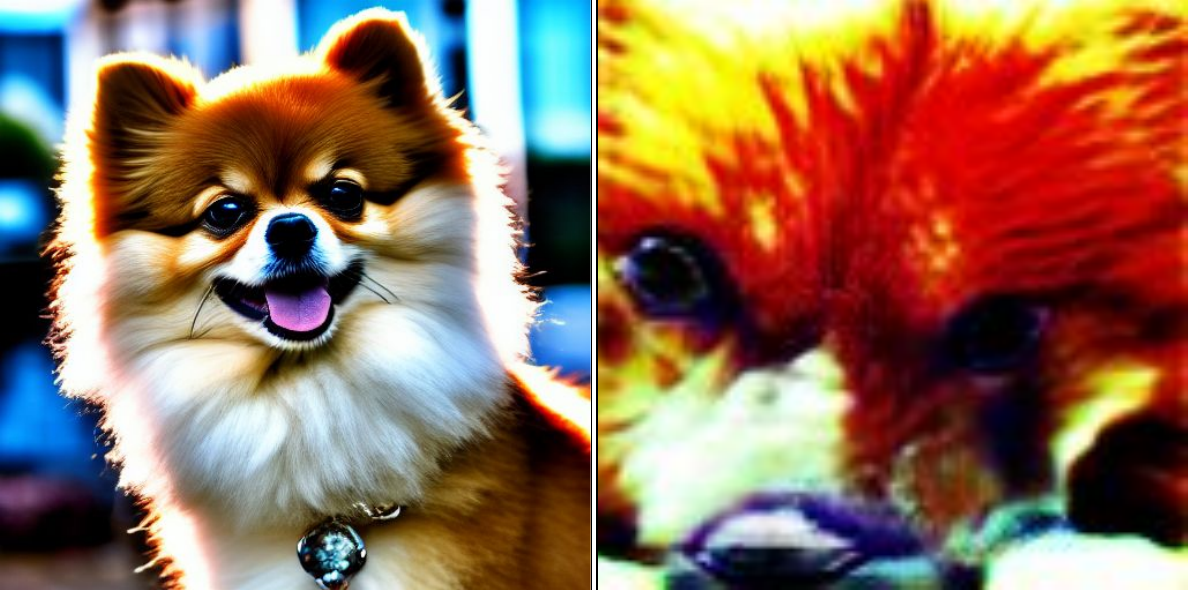}
    & \includegraphics[width=0.33\linewidth]{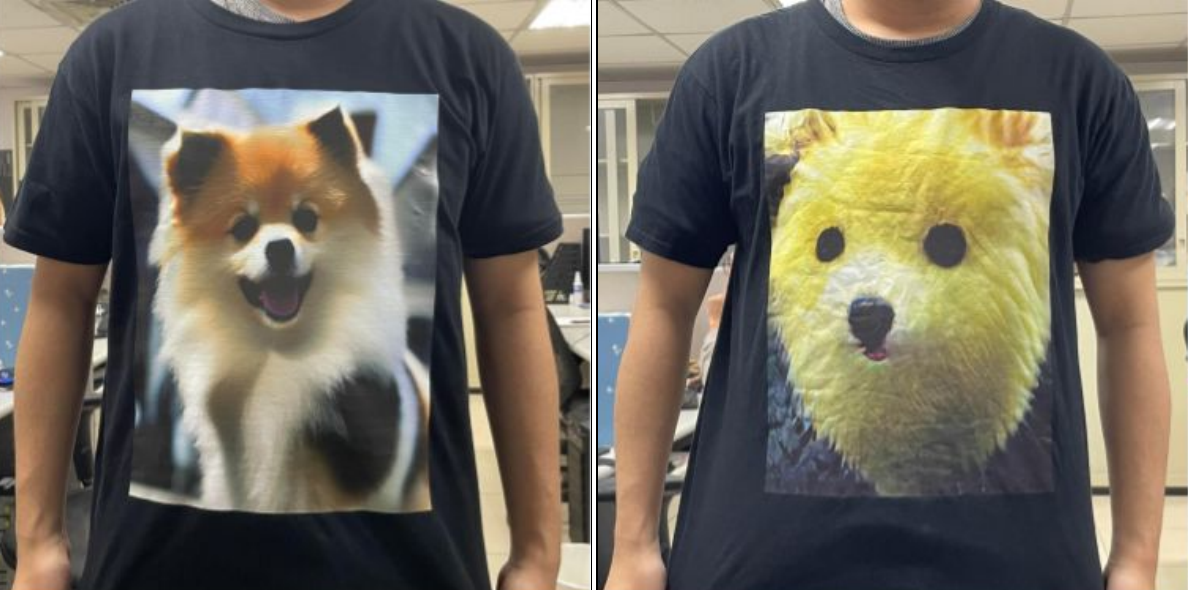}
    & \includegraphics[width=0.33\linewidth]{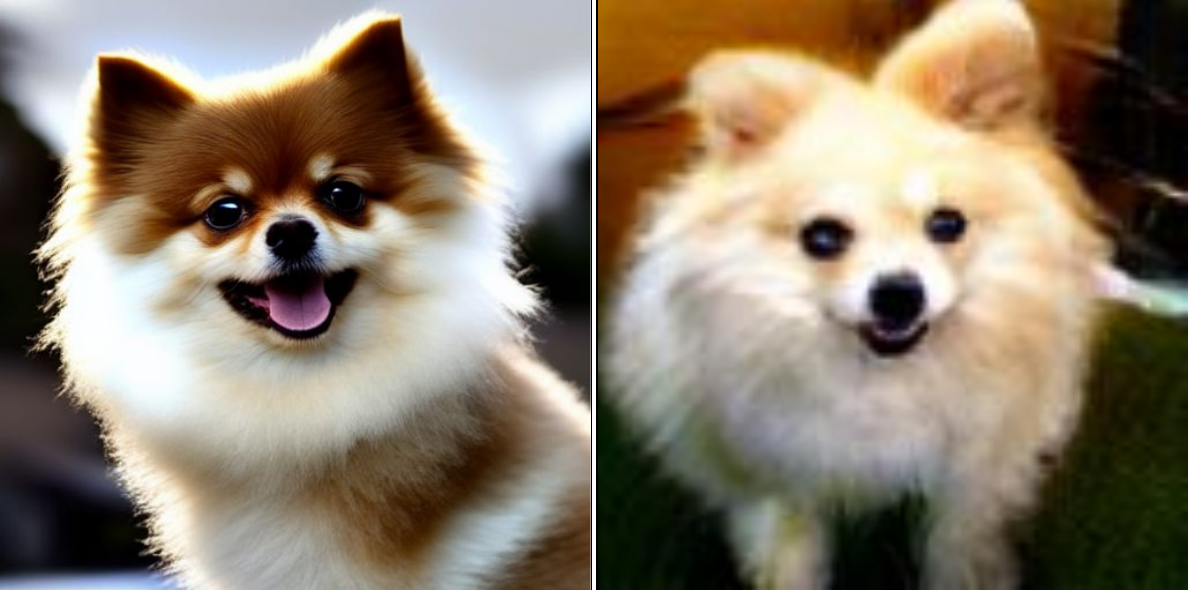} \\
    \includegraphics[width=0.33\linewidth]{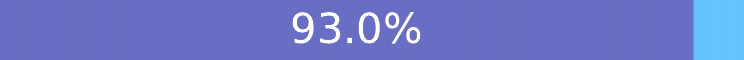}
    & \includegraphics[width=0.33\linewidth]{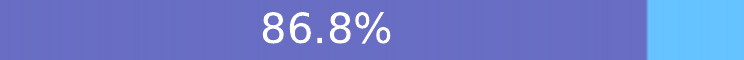}
    & \includegraphics[width=0.33\linewidth]{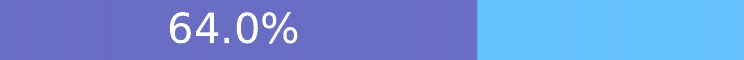}\\
    \bottomrule
\end{tabular*}
\end{table*}

We further conduct a comparison between our adversarial patches and the ones from previous works \cite{NPAP,UPC,legitimate,thys2019fooling}, as well as real images, in terms of naturalness. In each question, the participants are asked to rank three patches from the most natural to the least. The results in Table \ref{tab:user-study} demonstrate that our patches outperform all the previous works. Moreover, some participants consider our generated patches to be more natural than real-world images.

\begin{table*}
\centering
\caption{\label{tab:user-study} \textbf{Comparison between our adversarial patches and previous works \cite{NPAP, UPC, legitimate,thys2019fooling}}, as well as real images, in terms of naturalness. From left to right histograms
({\color[HTML]{696EC4} \textbf{purple}},
{\color[HTML]{66C3FF} \textbf{blue}}, and
{\color[HTML]{7CBE44} \textbf{green}})
represent the rankings of naturalness votes, from the
{\color[HTML]{696EC4} \textbf{most}}
natural to the
{\color[HTML]{7CBE44} \textbf{least}}.}
\begin{tabular*}{\linewidth}{@{\extracolsep{\fill}} cccccc}
   \toprule
    \multicolumn{3}{c}{Ensemble} & \multicolumn{3}{c}{YOLOv2} \\
    \cmidrule(lr){1-3}\cmidrule(lr){4-6}

    \includegraphics[width=0.15\linewidth]{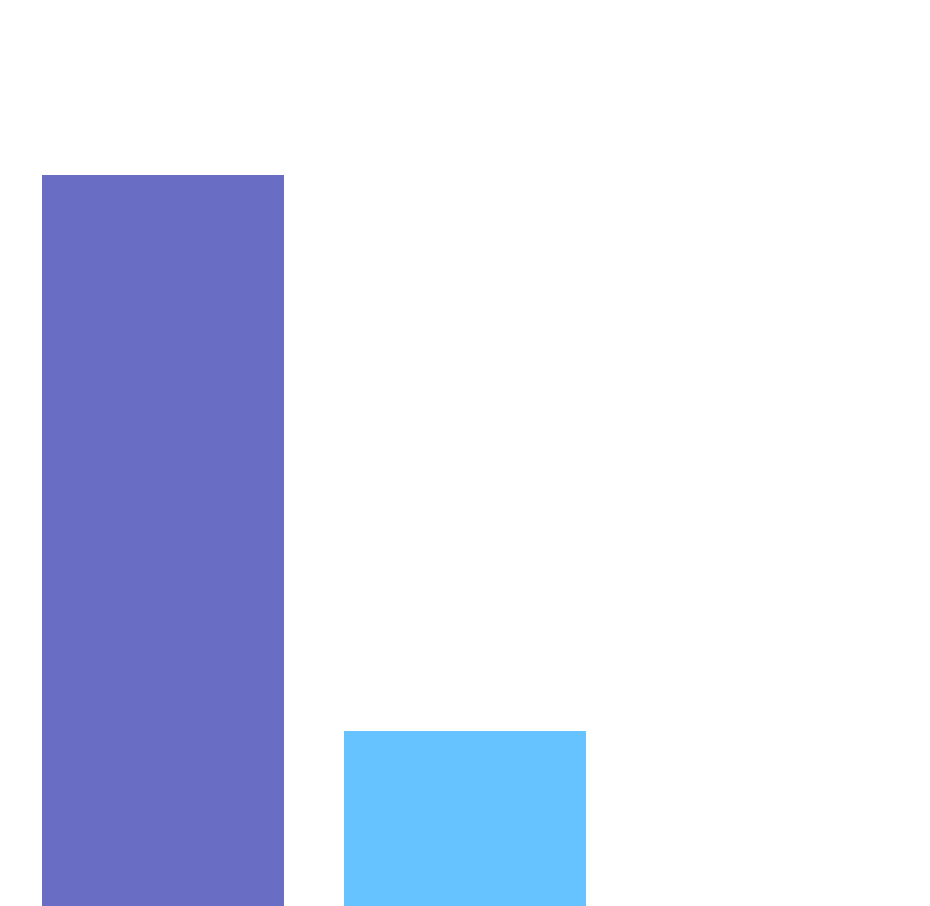}
    & \includegraphics[width=0.15\linewidth]{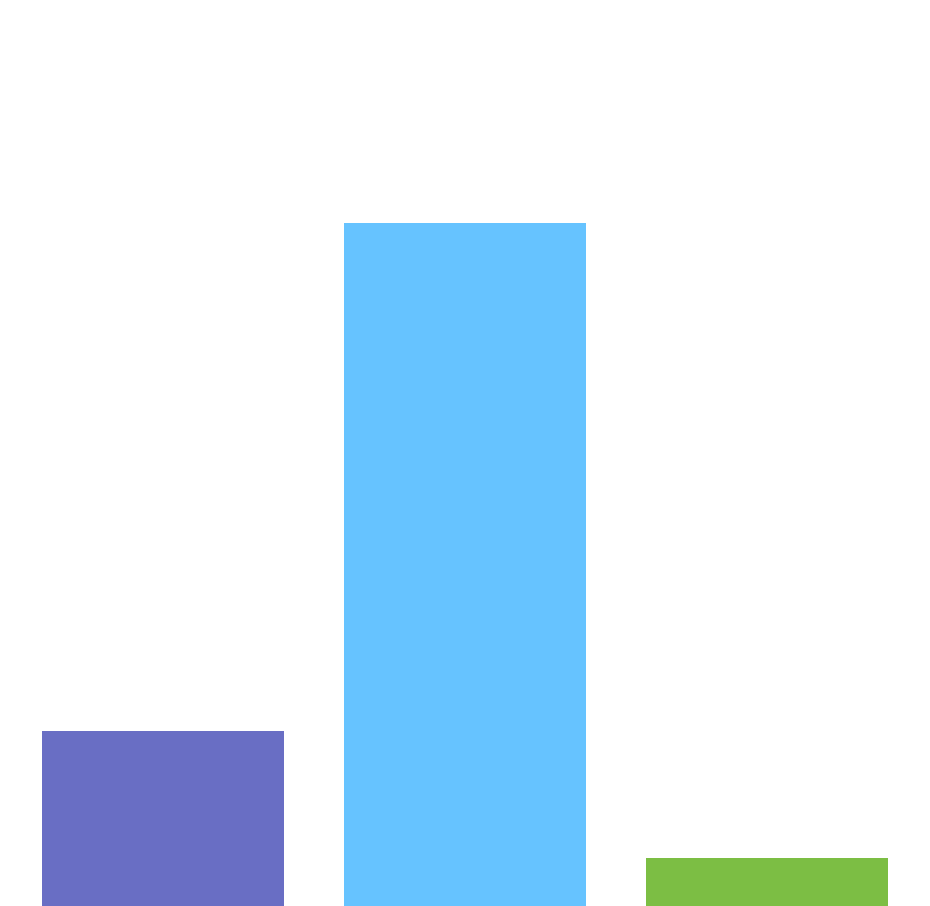}
    & \includegraphics[width=0.15\linewidth]{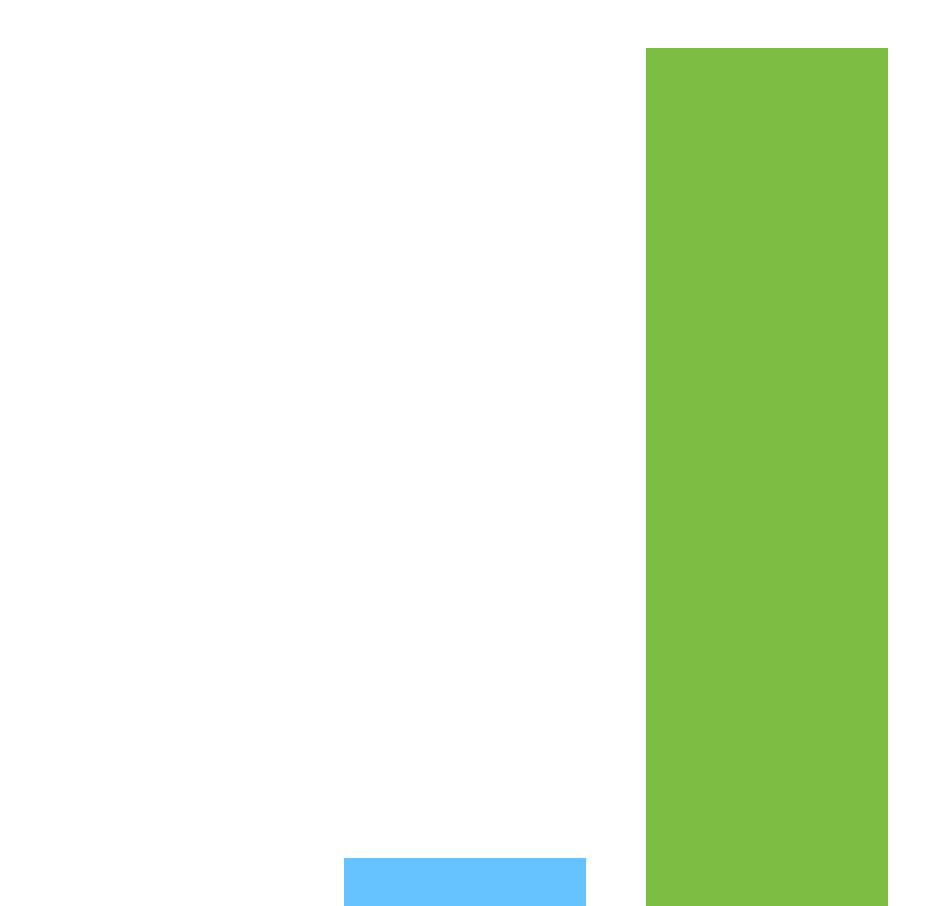}
    
    & \includegraphics[width=0.15\linewidth]{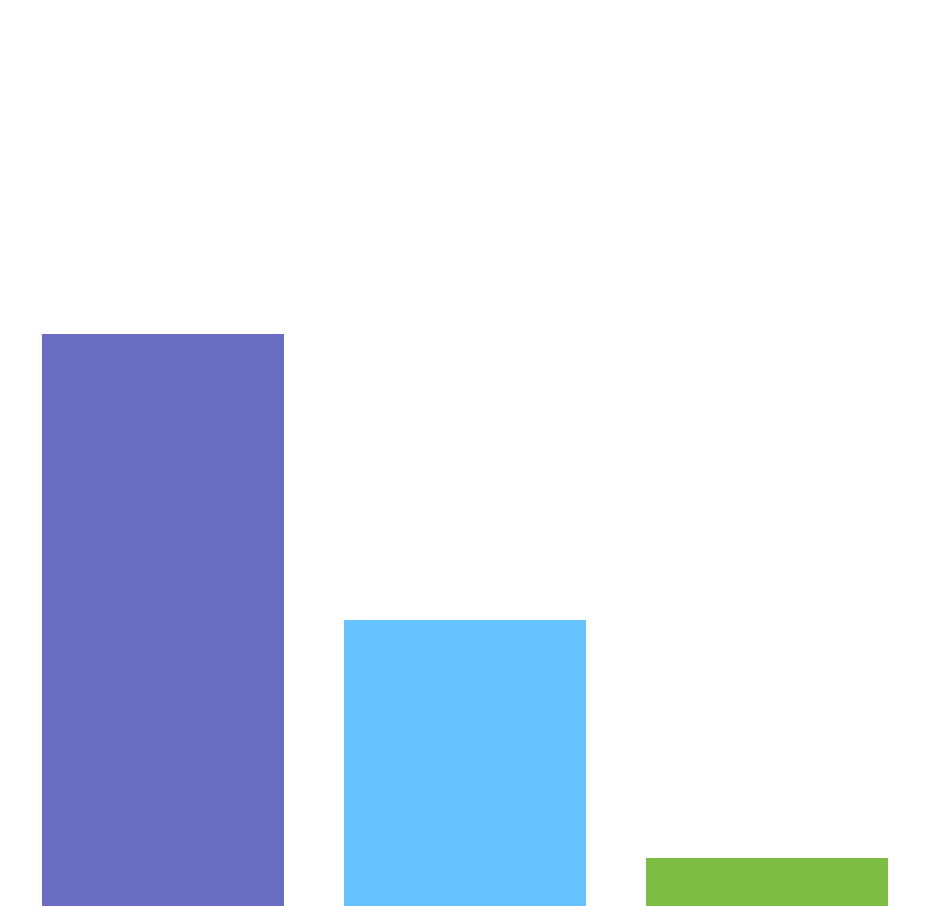}
    & \includegraphics[width=0.15\linewidth]{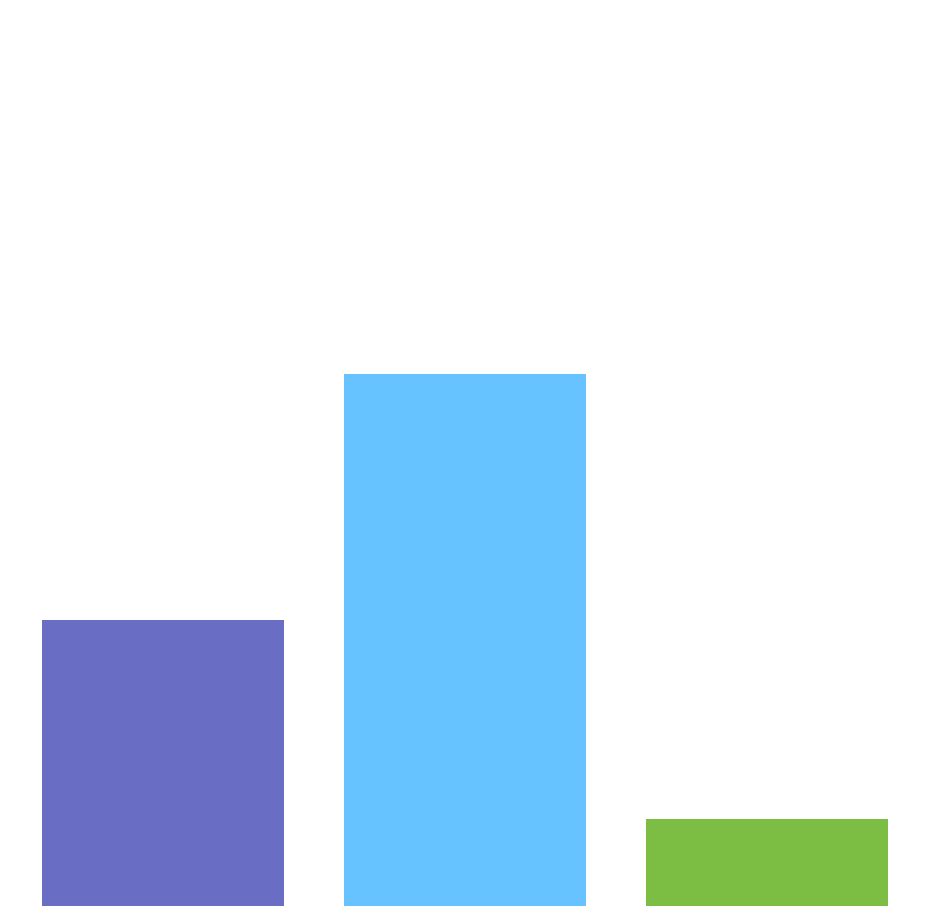}
    & \includegraphics[width=0.15\linewidth]{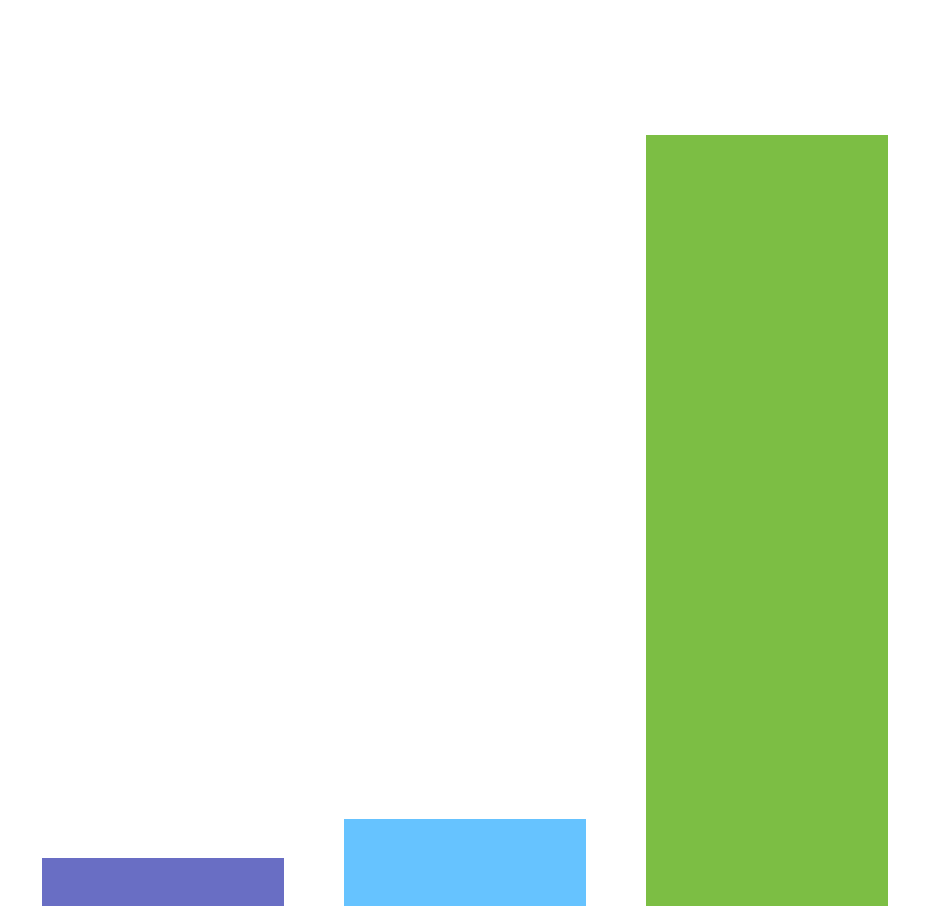} \\

    \includegraphics[width=0.15\linewidth]{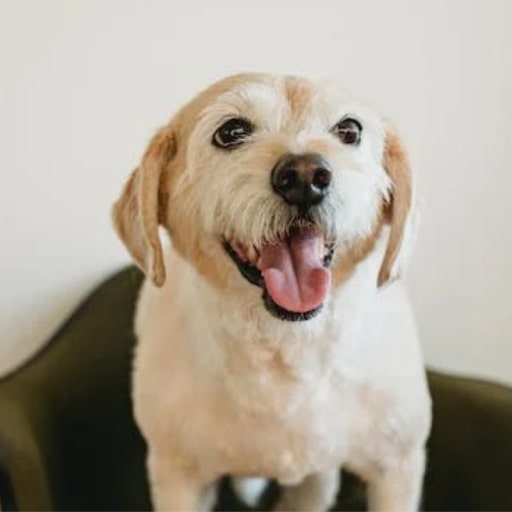}
    & \includegraphics[width=0.15\linewidth]{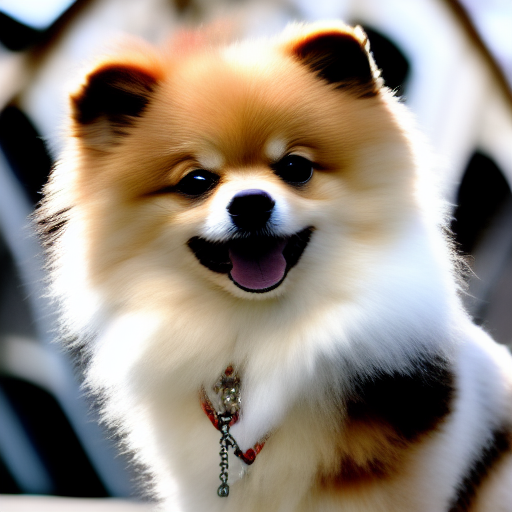}
    & \includegraphics[width=0.15\linewidth]{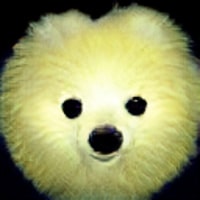}
    
    & \includegraphics[width=0.15\linewidth]{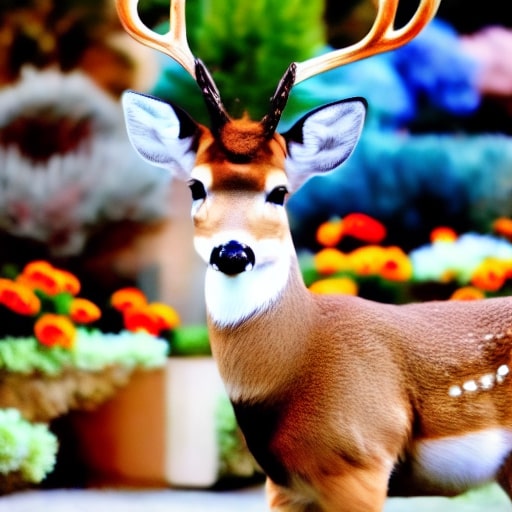}
    & \includegraphics[width=0.15\linewidth]{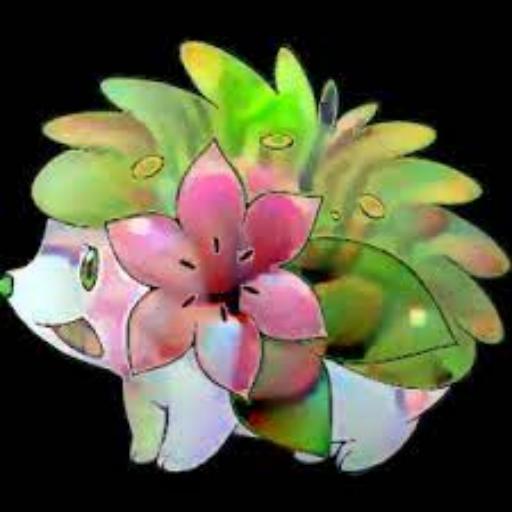}
    & \includegraphics[width=0.15\linewidth]{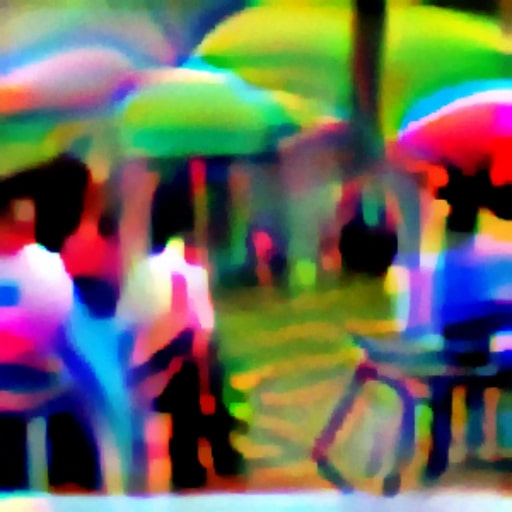} \\

    \makecell{Real-world \\image $^*$}
    & {Ours}
    & \makecell{Hu \etal \\\cite{NPAP}}
    
    & {Ours}
    & \makecell{Tan \etal \\\cite{legitimate}}
    & \makecell{Thys \etal \\\cite{thys2019fooling}} \\

    \toprule
    \multicolumn{3}{c}{FasterRCNN} & \multicolumn{3}{c}{YOLOv4-tiny} \\
    \cmidrule(lr){1-3}\cmidrule(lr){4-6}

    \includegraphics[width=0.15\linewidth]{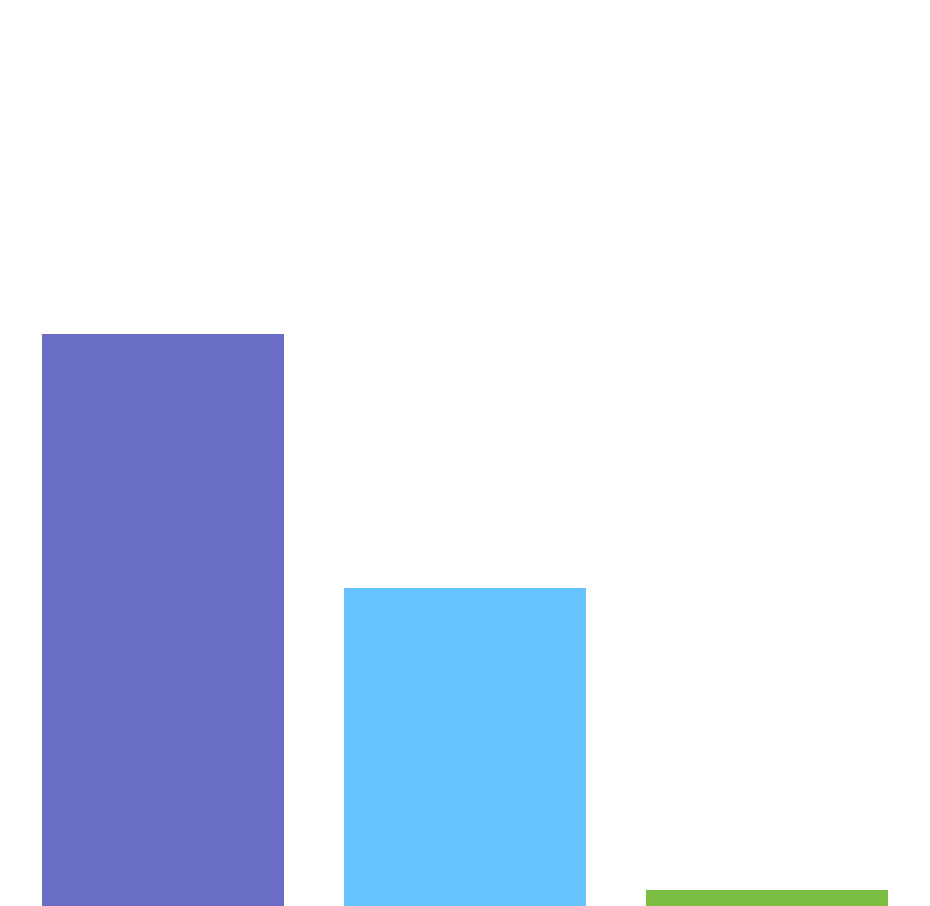}
    & \includegraphics[width=0.15\linewidth]{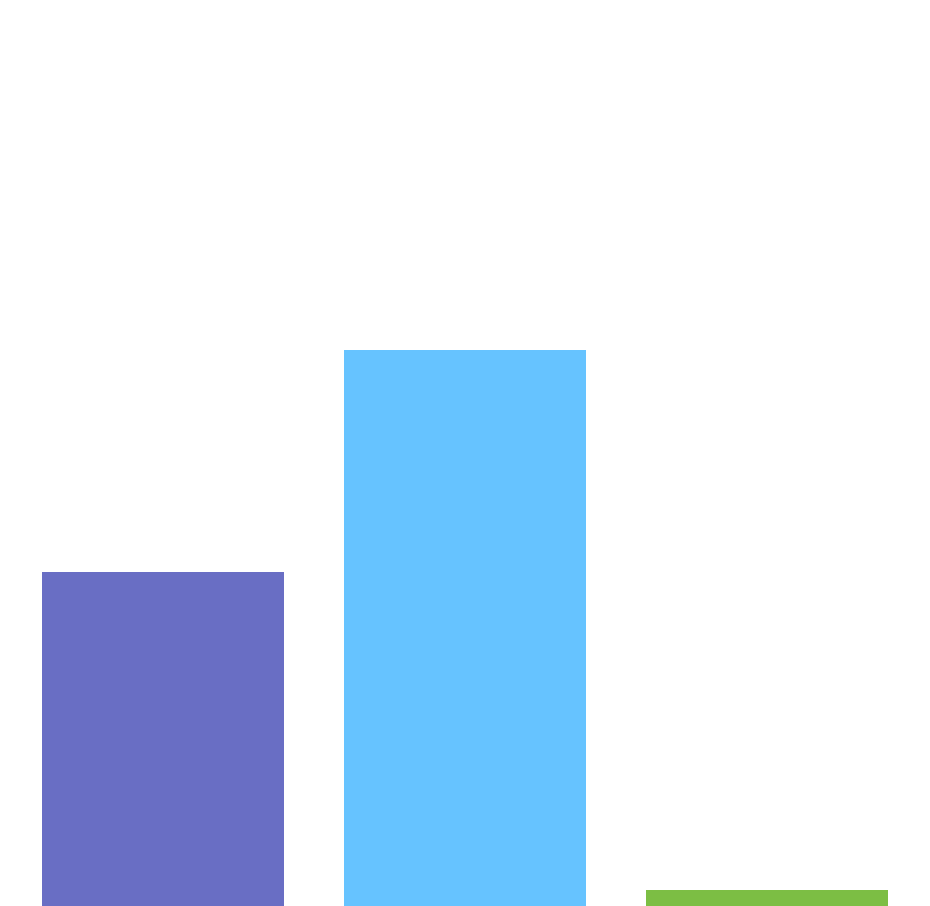}
    & \includegraphics[width=0.15\linewidth]{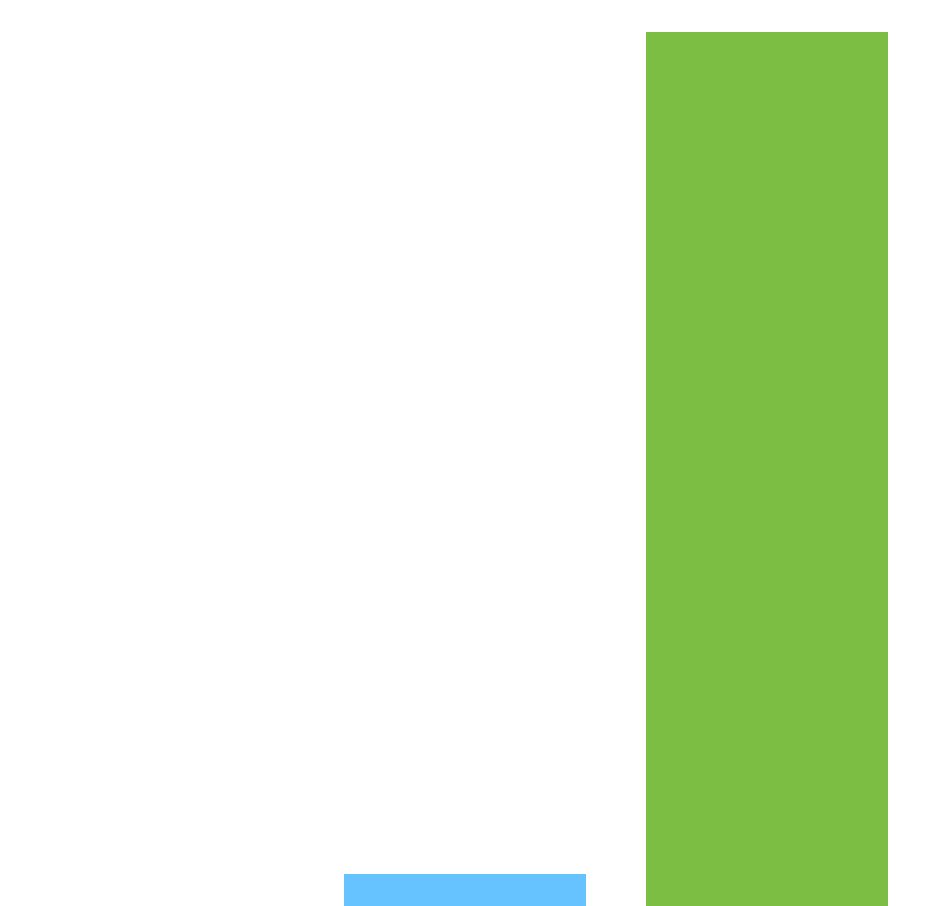}
    
    & \includegraphics[width=0.15\linewidth]{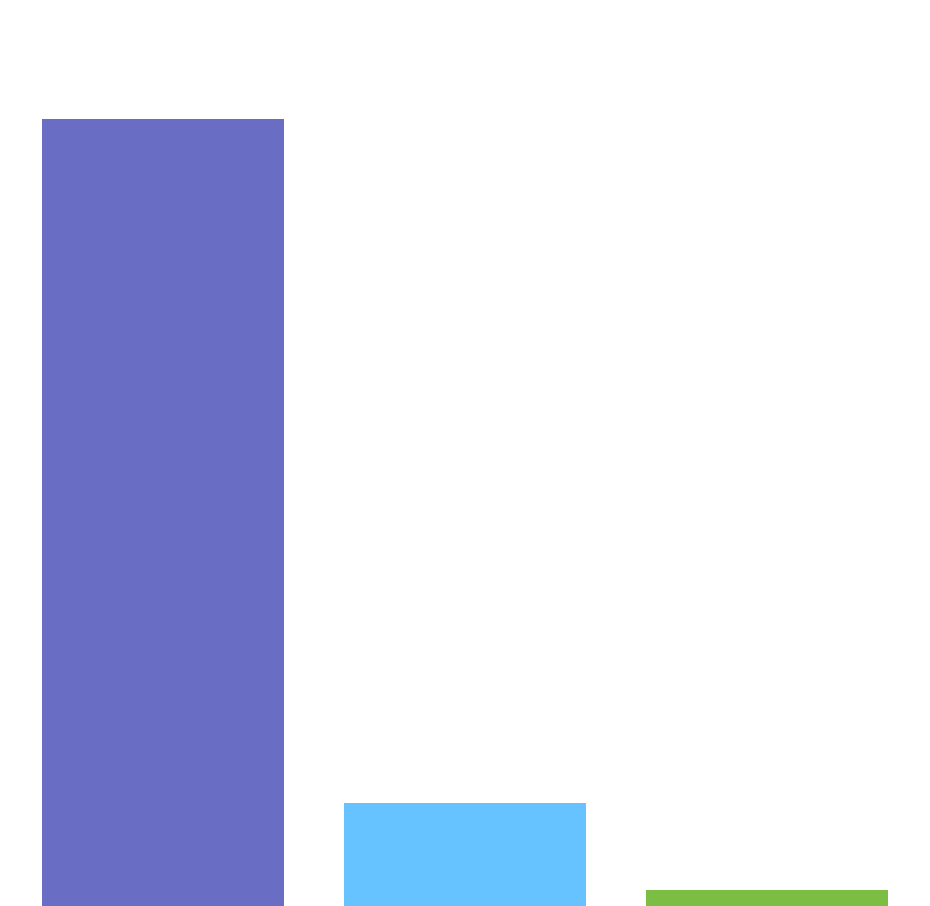}
    & \includegraphics[width=0.15\linewidth]{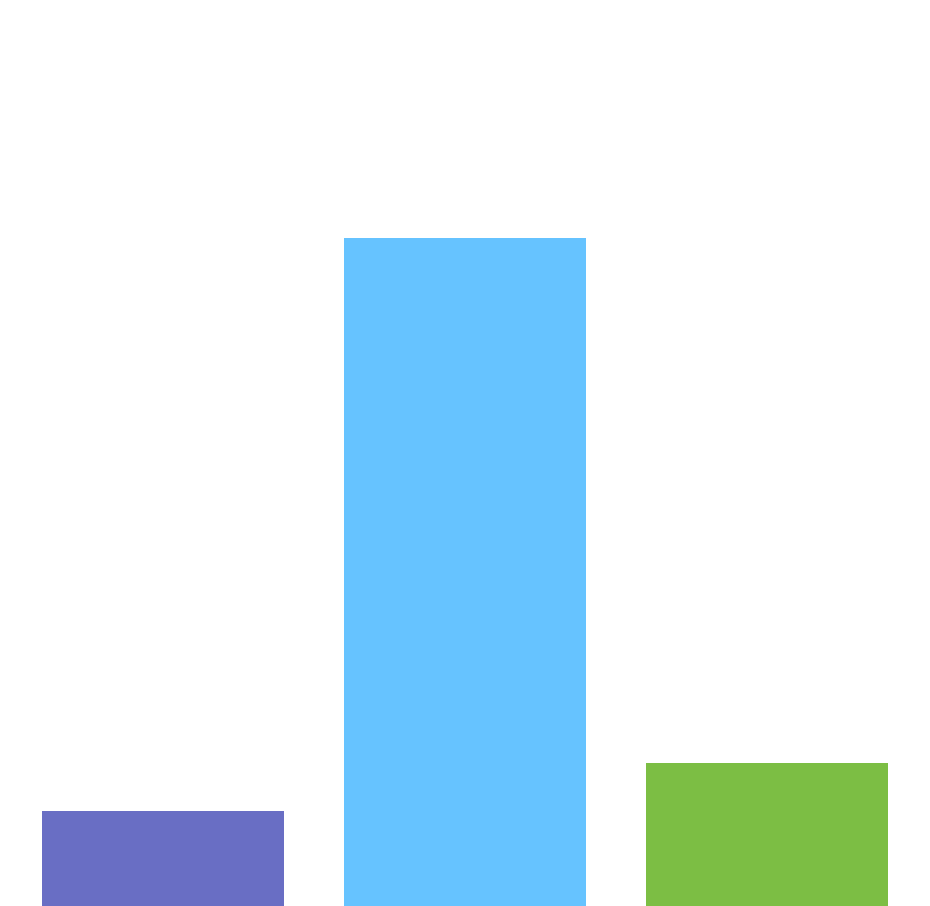}
    & \includegraphics[width=0.15\linewidth]{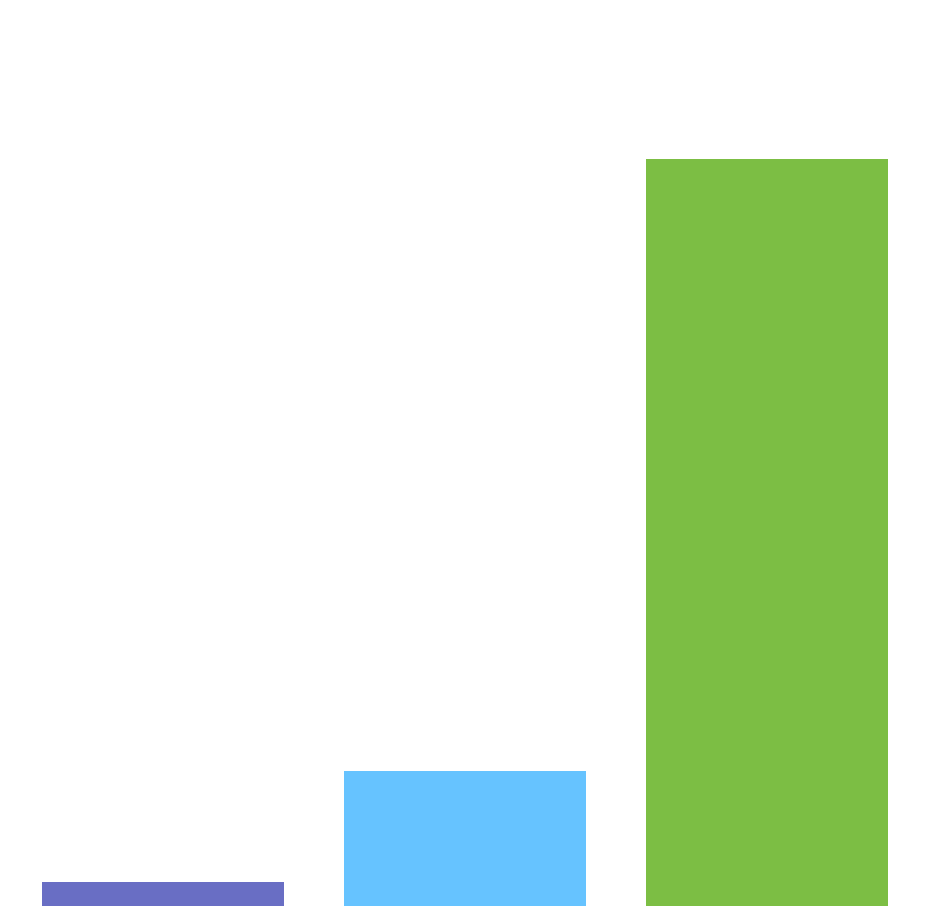} \\

    \includegraphics[width=0.15\linewidth]{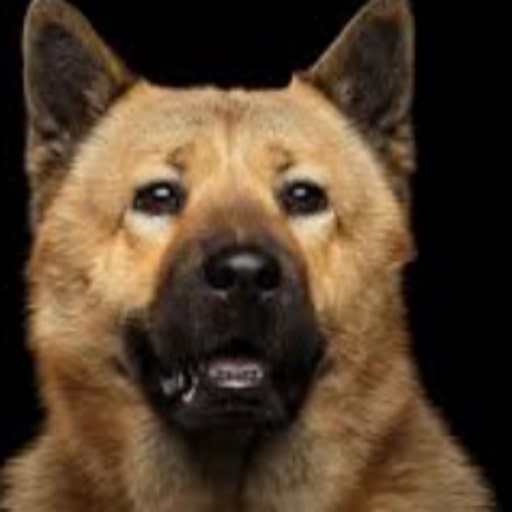} 
    & \includegraphics[width=0.15\linewidth]{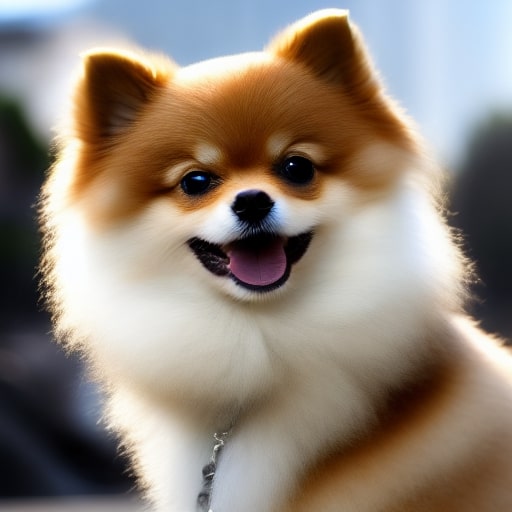}
    & \includegraphics[width=0.15\linewidth]{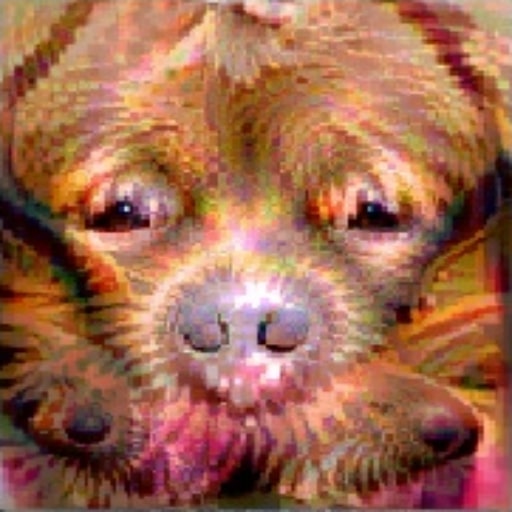}
    & \includegraphics[width=0.15\linewidth]{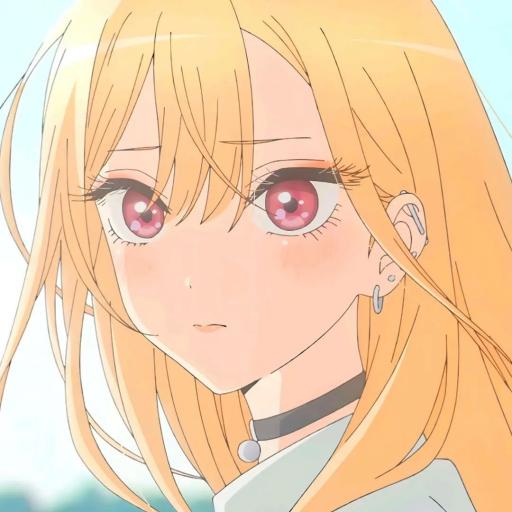}
    & \includegraphics[width=0.15\linewidth]{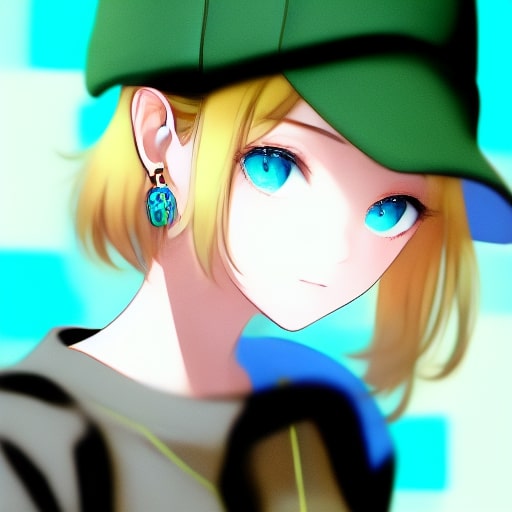}
    & \includegraphics[width=0.15\linewidth]{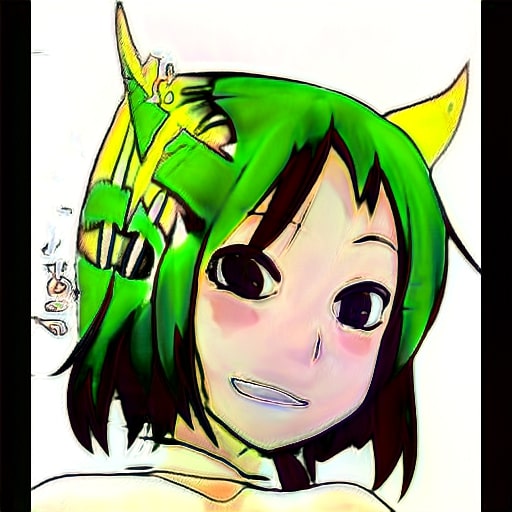} \\
    
    \makecell{Real-world  \\image $^\dagger$}
    & {Ours}
    & \makecell{Huang \etal \\\cite{UPC}}
    
    & \makecell{Real-world  \\image $^\ddagger$}
    & {Ours}
    & \makecell{Hu \etal \\\cite{NPAP}} \\
    \bottomrule
\end{tabular*}
\begin{tabular*}{\linewidth}{@{\extracolsep{\fill}} l p{15cm}}
     Image credit & $^*$ \url{https://www.pexels.com/photo/adorable-purebred-puppy-with-tongue-out-on-chair-5255202} \\
     & $^\dagger$ \url{https://dogtime.com/dog-breeds/akita-chow} \\
     & $^\ddagger$ \url{https://www.pinterest.com/pin/217791331971478398} \\
     \\
\end{tabular*}
\end{table*}

\section{Classifier-free guidance scale}
Ho and Salimans \cite{ho2021classifierfree} proposed classifier-free guidance (CFG) to control the output content of the diffusion model. They randomly mask out the conditional input during training the denoising U-Net, making the U-Net capable of both unconditional and conditional denoising. During sampling, a CFG weight $w$ controls the amount of extrapolation from the unconditional noise prediction towards the conditional one. We sweep over a range of $w$ (i.e., from the values of $\{1,2,...,20\}$) to see if higher text condition guidance helps boost the naturalness of the generated patches, as shown in Figure \ref{fig:cfg}.
\begin{figure*}
    \begin{tikzpicture}[scale=0.7525]
        \node at (-0.85,0) {};
        \foreach \i in {1,...,20}{
            \node at (\i,0) {\includegraphics[width=0.04\linewidth]{supple_fig/scale//\i.jpg}};
        }
    \end{tikzpicture}
    \vskip -0.15cm
    \begin{tikzpicture}
        \pgfplotsset{
            every tick label/.append style={font=\small},
            yticklabel={\pgfmathprintnumber[assume math mode=true]{\tick}}
        }
        \begin{axis}[
            xmin=0.5, xmax=20.5,
            axis y line*=left,
            width=.95\linewidth,
            height=6cm,
            xlabel={$w$},
            ylabel={\small mAP(\%)},
            y label style={at={(axis description cs:0.04,.5)},anchor=south},
            y dir=reverse
        ]
            \addplot[smooth, mark=triangle, blue] table [x=scale, y=mAP, col sep=comma] {supple_fig/scale/stat.csv};
            \label{plot:mAP_2}
        \end{axis}
        
        \begin{axis}[
            xmin=0.5, xmax=20.5,
            axis y line*=right,
            width=.95\linewidth,
            height=6cm,
            ylabel={\small CLIP Sim.(\%)},
            legend columns=-1,
            y label style={at={(axis description cs:1.16,.5)},anchor=south},
            legend style={at={(0.625,0.2)}}
        ]
            \addlegendimage{/pgfplots/refstyle=plot:mAP_2}\addlegendentry{\footnotesize mAP}
            \addplot[smooth, mark=square, red] table [x=scale, y=clip_sim, col sep=comma] {supple_fig/scale/stat.csv};
            \addlegendentry{\footnotesize CLIP Sim.}
        \end{axis}
    \end{tikzpicture}
    \caption{\textbf{Effects of the classifier-free diffusion guidance scale.}}
    \label{fig:cfg}
\end{figure*}

\section{Different pretrained models}
Many derivative pretrained weights have been released by fine-tuning the original Stable Diffusion \cite{LDM} under various conditions and datasets. In addition to Stable Diffusion v1.4, we have used in the main text, we also run experiments on, Latent Diffusion Model, Waifu Diffusion, and the recently released Stable Diffusion v2. Table \ref{tab:different-weights}  presents the results obtained using different pretrained weights, demonstrating the versatility of our method across various diffusion model weights.

\begin{table*}
\centering
\caption{\label{tab:different-weights} \textbf{Attack performance (mAP\%) on YOLOv4-tiny with other different pretrained weights.}}
\begin{tabular*}{\linewidth}{@{\extracolsep{\fill}} cccccc}
   \toprule
    \multicolumn{2}{c}{Latent Diffusion Model} & \multicolumn{2}{c}{Waifu Diffusion} & \multicolumn{2}{c}{Stable Diffusion v2} \\
    \cmidrule(lr){1-2}\cmidrule(lr){3-4}\cmidrule(lr){5-6}
    
    \includegraphics[width=0.15\linewidth]{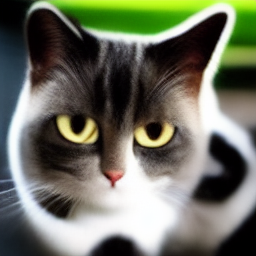}
    & \includegraphics[width=0.15\linewidth]{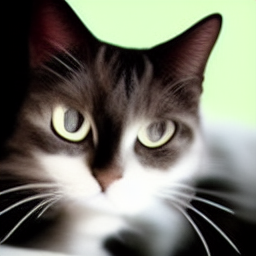}
    & \includegraphics[width=0.15\linewidth]{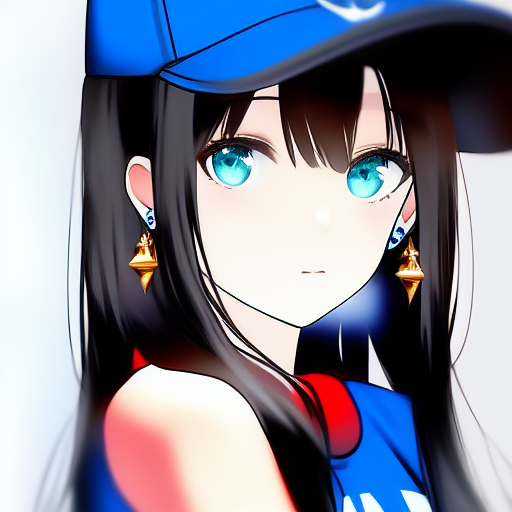}
    & \includegraphics[width=0.15\linewidth]{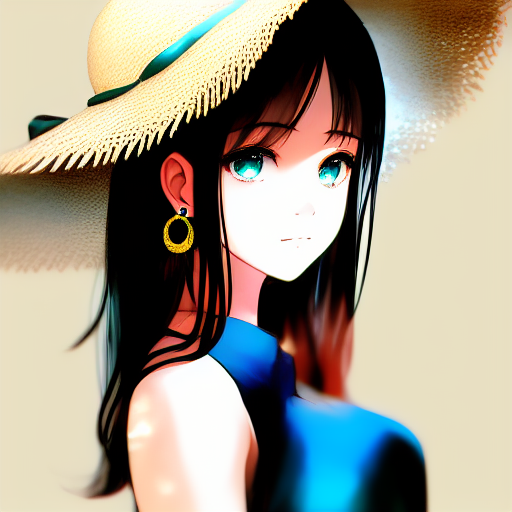}
    & \includegraphics[width=0.15\linewidth]{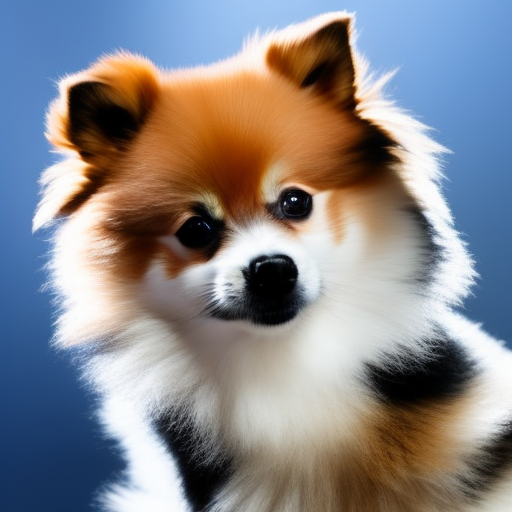}
    & \includegraphics[width=0.15\linewidth]{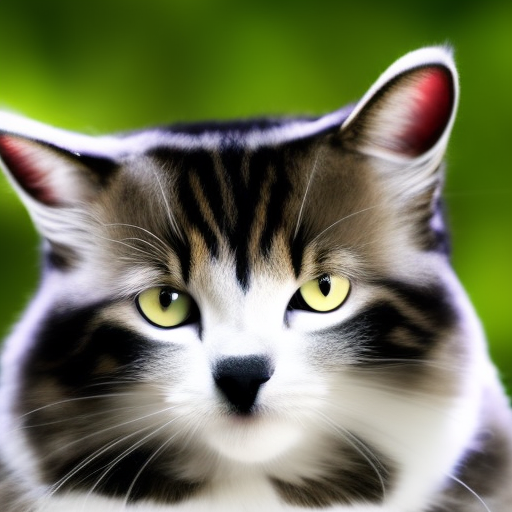}
     \\
    12.47 & 18.29 & 25.96 & 28.49 & 14.75 & 17.60 \\
\bottomrule
\end{tabular*}
\end{table*}

\end{document}